\documentclass{article}

\usepackage{microtype}
\usepackage{graphicx}
\usepackage{booktabs} 

\usepackage{hyperref}

\usepackage[accepted]{icml2023}

\usepackage{amsmath}
\usepackage{amssymb}
\usepackage{mathtools}
\usepackage{graphicx}
\usepackage{amsthm}
\usepackage{booktabs}

\usepackage[T1]{fontenc}
\usepackage{etoc}
\usepackage[most]{tcolorbox}
\usepackage{multirow}
\usepackage{float}
\usepackage{subfig}
\usepackage{comment}
\usepackage{multicol}
\usepackage{appendix}
\usepackage{bbm}
\usepackage[inkscapelatex=false]{svg}
\usepackage{bbding}
\usepackage{framed}
\usepackage{setspace}
\usepackage{pifont}
\newcommand{\cmark}{\ding{51}}
\newcommand{\xmark}{\ding{55}}
\usepackage{xcolor}         
\definecolor{mycolor1}{rgb}{0.82,0.41,0.12}
\definecolor{mycolor2}{rgb}{0.0,0.51,0.22}
\definecolor{mycolor3}{rgb}{0.98,0.75,0.75}
\definecolor{mycolor4}{rgb}{0.2,0.7,0.2}

\usepackage{textcomp,booktabs}

\usepackage[capitalize,noabbrev]{cleveref}

\theoremstyle{plain}
\newtheorem{theorem}{Theorem}[section]
\newtheorem{proposition}[theorem]{Proposition}
\newtheorem{lemma}[theorem]{Lemma}

\theoremstyle{definition}

\theoremstyle{remark}

\usepackage[textsize=tiny]{todonotes}

\icmltitlerunning{Calibrating Multimodal Learning}

\begin{document}

\twocolumn[
\icmltitle{Calibrating Multimodal Learning}



\icmlsetsymbol{equal}{*}

\begin{icmlauthorlist}
\icmlauthor{Huan Ma}{equal,tju,comp}
\icmlauthor{Qingyang Zhang}{equal,tju}
\icmlauthor{Changqing Zhang}{tju,keylab} \\
\icmlauthor{Bingzhe Wu}{comp}
\icmlauthor{Huazhu Fu}{astar}
\icmlauthor{Joey Tianyi Zhou}{astar,cfar}
\icmlauthor{Qinghua Hu}{tju,keylab}

\end{icmlauthorlist}

\icmlaffiliation{tju}{College of Intelligence and Computing, Tianjin University, Tianjin, China}
\icmlaffiliation{comp}{AI Lab, Tencent, Shenzhen, China}
\icmlaffiliation{astar}{Institute of High Performance Computing (IHPC), Agency for Science, Technology and Research (A*STAR), Singapore}
\icmlaffiliation{cfar}{Centre for Frontier AI Research (CFAR), Agency for Science, Technology and Research (A*STAR), Singapore}
 \icmlaffiliation{keylab}{Tianjin Key Lab of Machine Learning, Tianjin, China}
\icmlcorrespondingauthor{Changqing Zhang}{zhanchangqing@tju.edu.cn}
\icmlcorrespondingauthor{Bingzhe Wu}{bingzhewu@tencent.com}

\icmlkeywords{Machine Learning, ICML}

\vskip 0.3in
]



\printAffiliationsAndNotice{\icmlEqualContribution} 

\begin{abstract}
Multimodal machine learning has achieved remarkable progress in a wide range of scenarios. However, the reliability of multimodal learning remains largely unexplored. In this paper, through extensive empirical studies, we identify current multimodal classification methods suffer from unreliable predictive confidence that tend to rely on partial modalities when estimating confidence. Specifically, we find that the confidence estimated by current models could even increase when some modalities are corrupted. To address the issue, we introduce an intuitive principle for multimodal learning, i.e., the confidence should not increase when one modality is removed. Accordingly, we propose a novel regularization technique, i.e., Calibrating Multimodal Learning (CML) regularization, to calibrate the predictive confidence of previous methods. This technique could be flexibly equipped by existing models and improve the performance in terms of confidence calibration, classification accuracy, and model robustness.
\end{abstract}

\etocdepthtag.toc{mtchapter}
\etocsettagdepth{mtchapter}{subsection}
\etocsettagdepth{mtappendix}{none}

\section{Introduction}

Multimodal data widely exist in real-world applications such as medical analysis~\cite{perrin2009multimodal}, social media~\cite{2019word}, and autonomous driving~\cite{Khodayari2010A}. To fully explore the potential value of each modality, multimodal learning emerges as a promising way to
train a machine learning (ML) model by integrating all available multimodal cues for further data analysis tasks. Numerous approaches have been proposed to build multimodal learning paradigms for various tasks~\cite{2019word,antol2015VQA,bagher2018multimodal,kishi2019Correlation}.
Despite above progresses, the reliability of current multimodal learning methods remains largely unexplored. In the setting of classification, one key aspect of the reliability is to build a high-quality confidence estimator~\cite{moon2020Confidence,2020address,guo2017On}, which can quantitatively characterize the probability that predictions will be correct. With such an estimator, further processing can be taken to improve the performance of the system (e.g., human assistance) when the predictive uncertainty is high. This is especially useful in high-stake scenarios~\cite{2018Noise,QaddoumReliable}.

\begin{figure*}[th]
    \centering
    \includegraphics[width=0.94\textwidth]{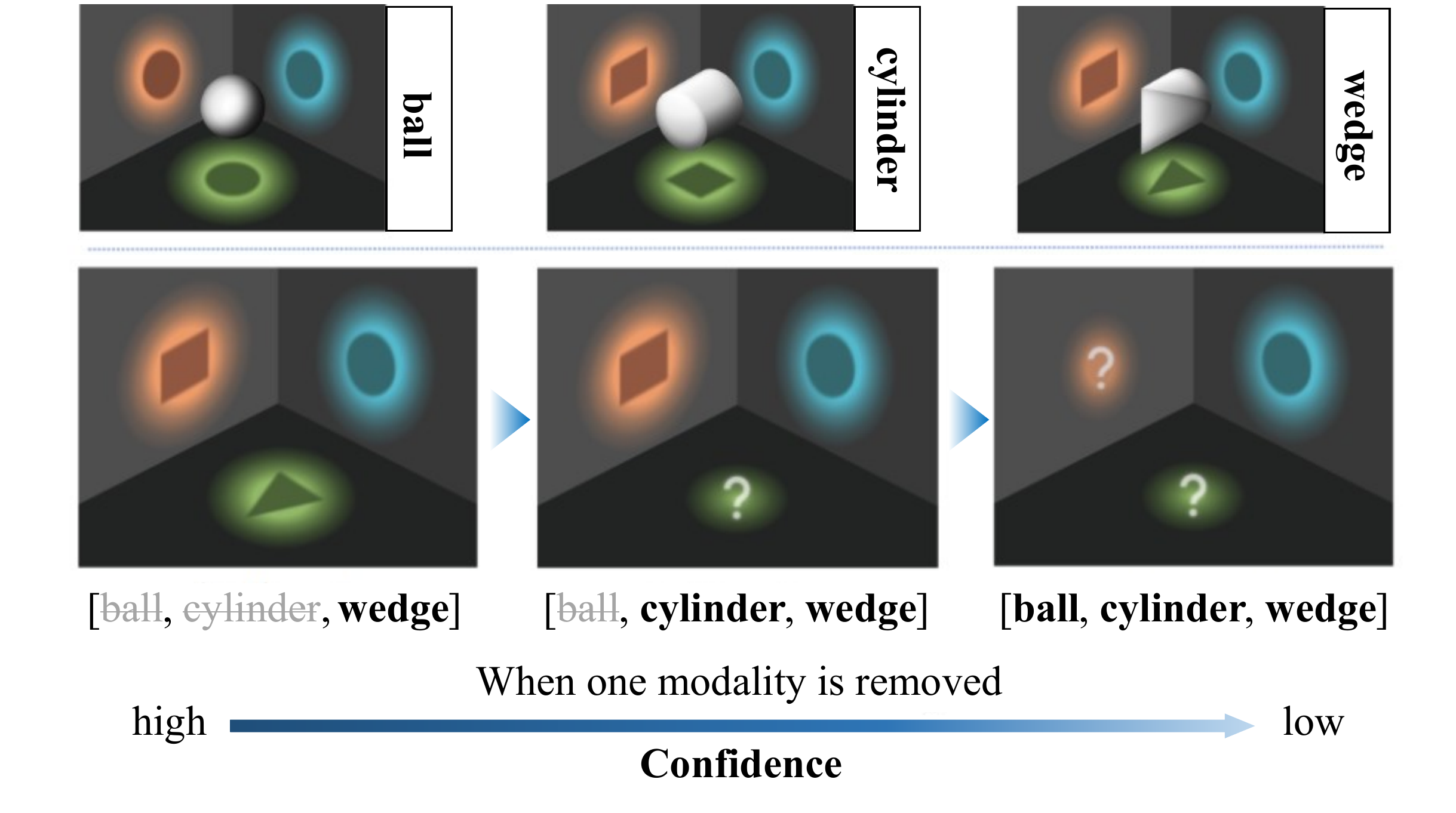}
    \caption{Motivation of calibrating multimodal learning. The confidence of an ideal multimodal classifier should decrease or at least not increase when one modality is removed  (even when the removed modality is noised, or it indicates the model takes noise as semantics and the model is not trustworthy).}
    \label{fig:coverfigure}
\end{figure*}

In the setting of multimodal learning, in addition to exact overall prediction confidence, the relationship between the confidence and the number of modalities should also be taken into concerns. Intuitively, the confidence of an ideal multimodal classifier should not increase when one modality is removed (for brevity, we initialize the question as ``one modality", and the same phenomenon is observed when removing more than one modality). An illustrative example of an ideal confidence estimator is shown in Fig.~\ref{fig:coverfigure}, where the confidence gradually decreases when the observed information becomes less comprehensive. However, we conduct extensive empirical studies on current methods and observe that when one modality is removed, the overall confidence estimated by them can even increase. This observation contradicts the common assumption of multimodal learning since modalities
are assumed to be predictive of the target for most multimodal learning tasks~\cite{wu2022Characterizing} and the principle ``\emph{the essence of information is to eliminate uncertainty (Shannon)}'' in informatics~\cite{soni2017mind,burgin2002essence}.
Intuitively, this implies that the models are more inclined to believe in a unique modality and is prone to be affected by this modality, which has also been shown in prior works \cite{wu2022Characterizing,wang2019What}. This further impairs the robustness of the learned models, i.e., the models are easy to be influenced when some modalities are corrupted, since the models can not make decisions according to a trustworthy confidence (probability) estimator.

A natural idea to address the above issue is to employ recent uncertainty calibration methods such as temperature scaling \cite{guo2017On} or Bayesian learning \cite{cobb2020Scaling,kar2020Hierarchical,foong2019theexpress}, which can build more accurate confidence estimation than the traditional training/inference manner.
However, these approaches do not explicitly consider the relationship between different modalities (i.e., they can only calibrate the overall confidence but can not calibrate the confidence of using different number of modalities) and thus still fail to achieve satisfactory performance in the multimodal learning setting. To address this issue, we propose a novel regularization technique called \textbf{C}alibrating \textbf{M}ultimodal \textbf{L}earning (CML) which enforces the consistency between prediction confidence and the number of modalities. The motivation of CML is based on a natural intuition, i.e., the prediction confidence should decrease (at least not increase) when one modality is removed, which could intrinsically improve the confidence calibration. Specifically, we propose a simple regularization term that enforces a model to learn an intuitive ranking relationship by adding a penalty for the samples whose predictive confidence will increase when one modality is removed. 
The main contributions of this paper are summarized as follows:
\begin{itemize}
\setlength{\itemsep}{0mm}
\setlength{\parskip}{0pt}
\item We conduct extensive empirical studies to show that most existing multimodal learning paradigms tend to be over-confident on partial modalities (different samples are over-confident on different modalities rather than all samples are over-confident on the same modalities), which implies that they fail to achieve trustworthy confidence estimation.

\item We introduce a measure to evaluate the reliability of the confidence estimation from the confidence ranking perspective, which can characterize whether a multimodal learning method can treat all modalities fairly.

\item We propose a regularization strategy to calibrate the confidence of various multimodal learning methods, and then conduct extensive experiments to show the superiority of our method in terms of the confidence calibration (Table~\ref{tab:disorder}), classification accuracy (Table~\ref{tab:acc}) and model robustness (Table~\ref{tab:noise}).

\end{itemize}
\vspace{-0.2cm}

\section{Related Work}
\textbf{Uncertainty estimation} provides a way for trustworthy prediction~\cite{abdar2020asurvy,chau2021bayes,slack2021reliable,singh2021fairness,ning2021uncertainty,zhang2021relative}. Uncertainty can be used as an indicator of whether the predictions given by models are prone to be wrong~\cite{Ritter2021sparse,wang2021online,zaidi2021neural,stadler2021graph,bai2021understanding,rahaman2021uncertainty,galil2021disrupting,upadhyay2021robustness}. Many uncertainty-based models have been proposed in the past decades, such as Bayesian neural networks~\cite{neal2012bayesian,mackay1992bayesian,denker1990transforming,kendall2017uncertainties}, Dropout~\cite{2017dropout}, Deep ensembles~\cite{balaji2018ensemble,havasi2020training}, and DUQ~\cite{van2020uncertainty} built upon RBF networks. \textbf{Prediction confidence}~\cite{sa2021reliable,wa2021on,pan2021on,luo2021no,xu2021towards,chung2021beyond,xiong2021uncertain} is always referred to in classification models, which expects the predicted class probability to be consistent with the empirical accuracy~\cite{qin2021improving,mind2021revisit,zhao2021calibrating,tian2021a,karandikar2021soft,jeong2021smoothmix}. Many methods focus on smoothing the prediction probabilities distribution, such as Label smoothing~\cite{M2019when}, focal loss~\cite{Mukhoti2020Calibrating}, TCP~\cite{2020address}and Temperature scaling (TS)~\cite{guo2017On}. More related researches please refer to Appendix~\ref{sec:app-related}.

\begin{figure*}[!t] 
  \centering
  \subfloat[MMTM]{
  \centering
  \includegraphics[width=0.31\textwidth]{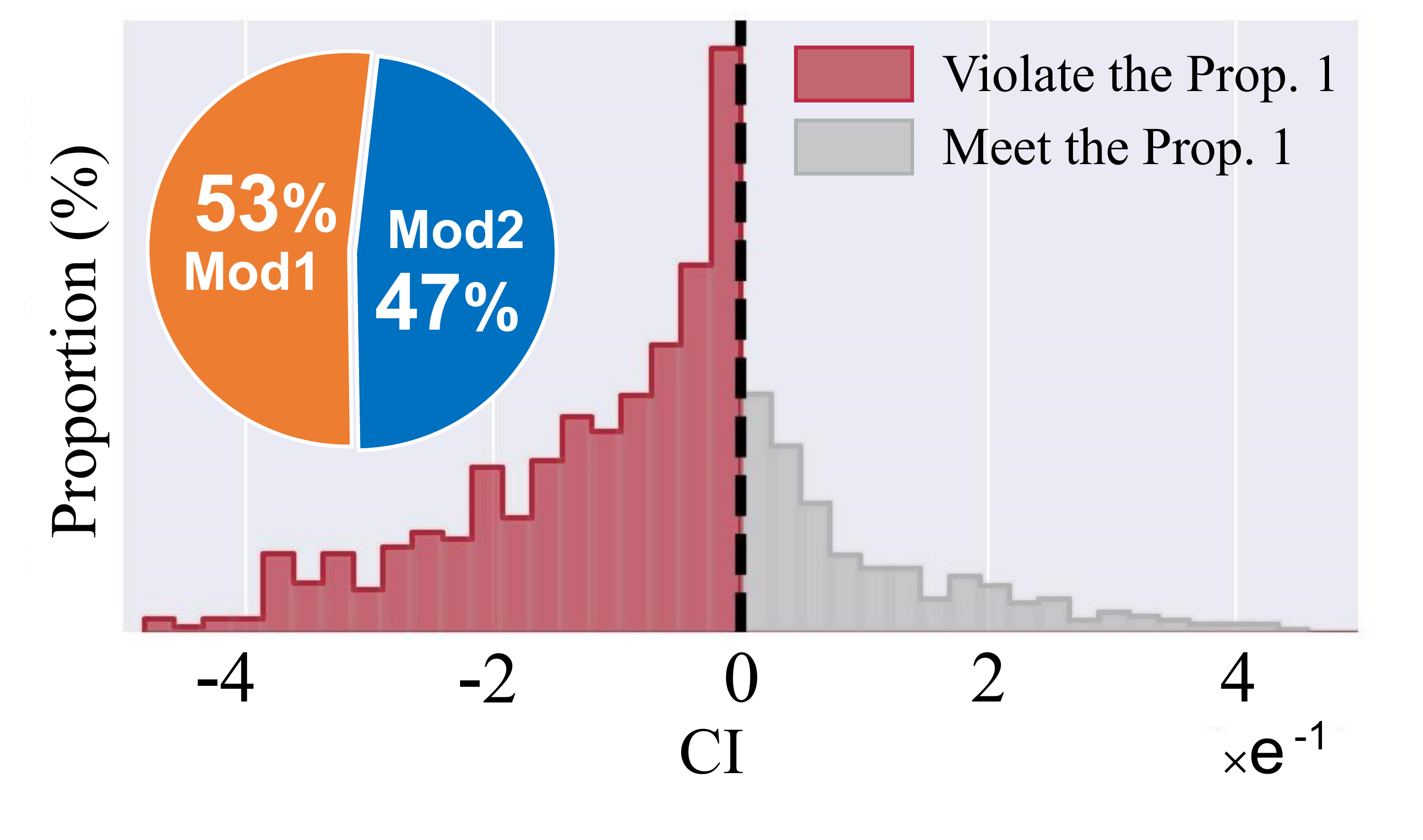}
  }
  \subfloat[CPM-Nets]{
  \centering
  \includegraphics[width=0.31\textwidth]{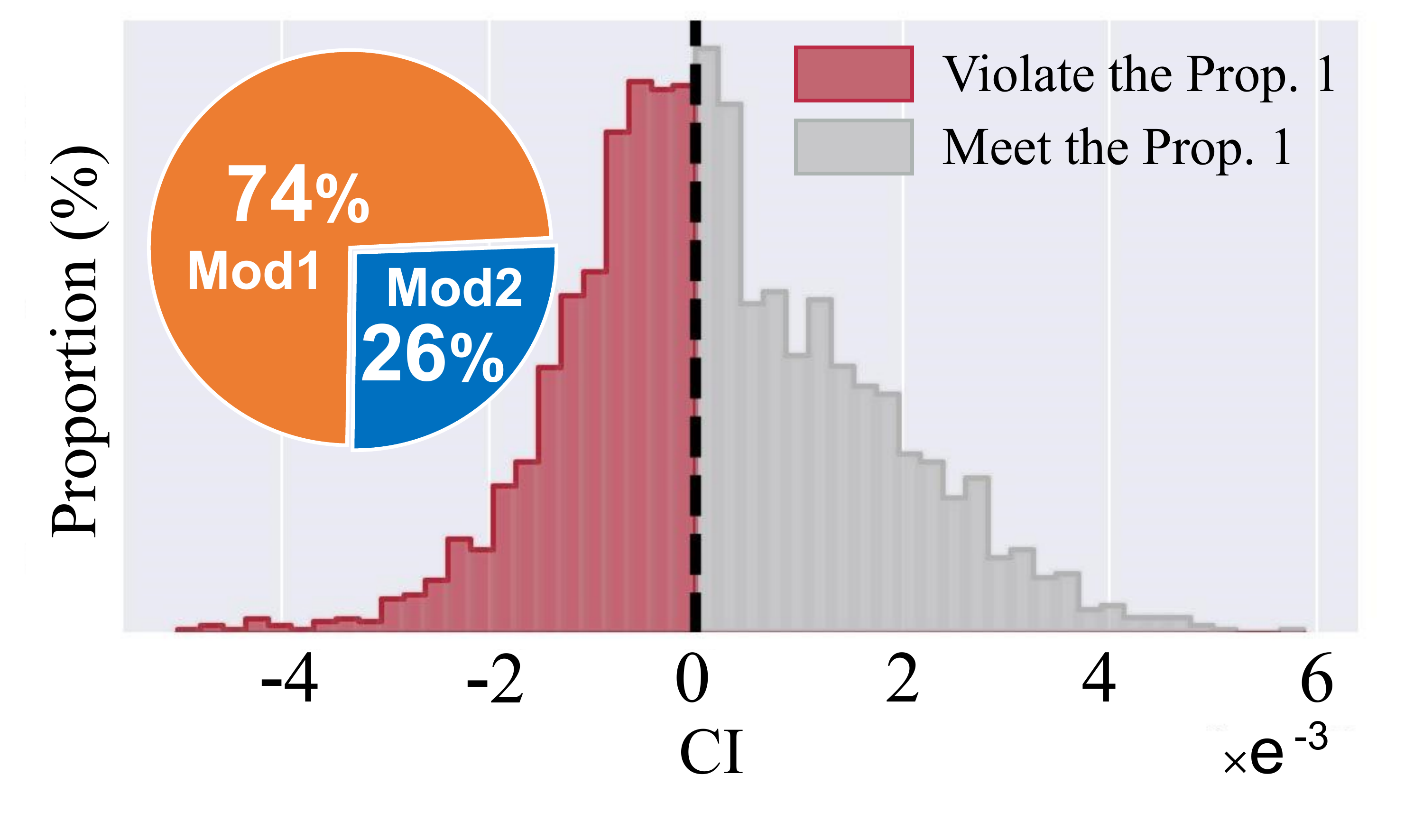}
  }
  \subfloat[MIWAE]{
  \centering
  \includegraphics[width=0.31\textwidth]{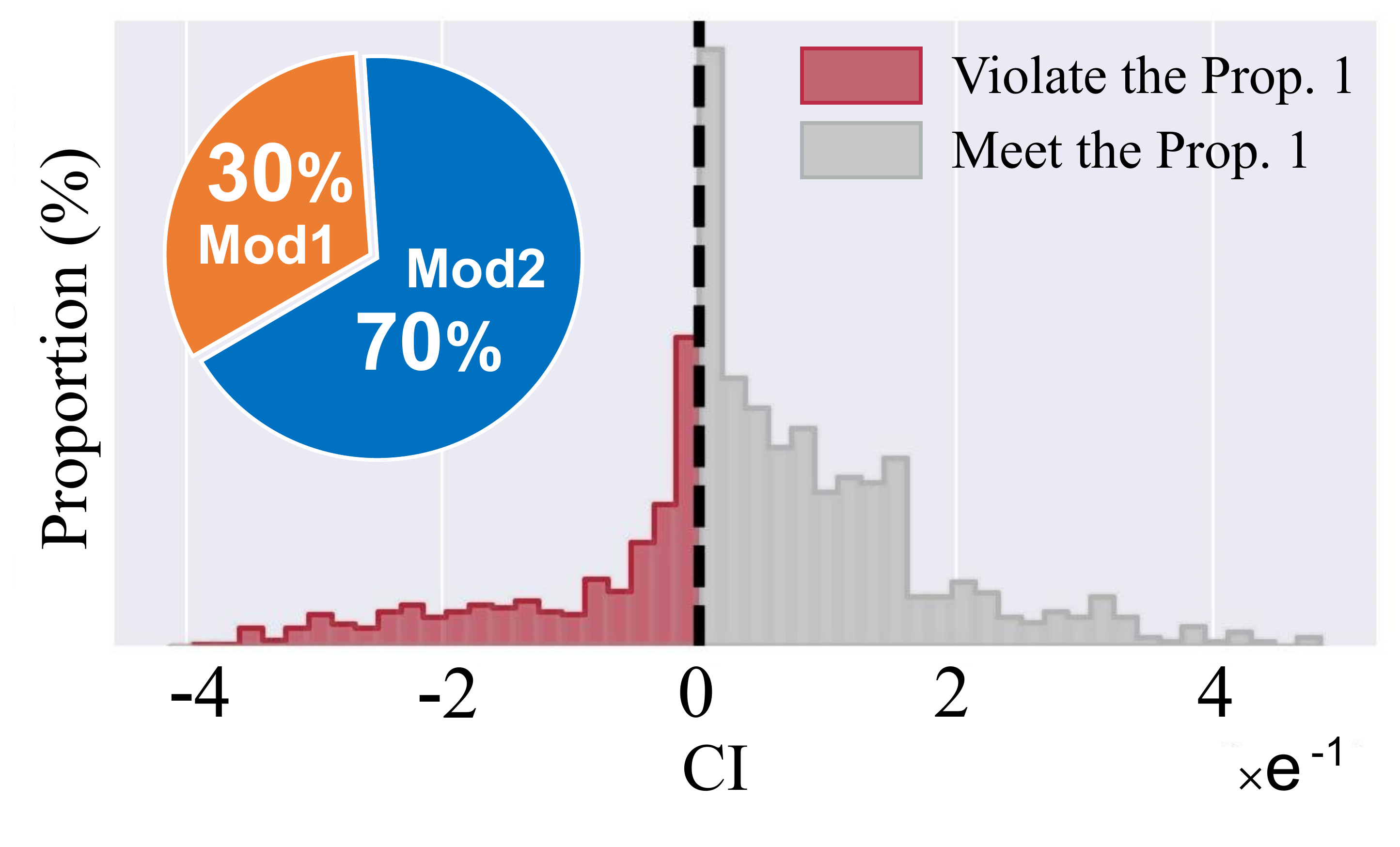}
  }
  
 \caption{Current methods \cite{wu2022Characterizing,zhang2019cpm,mattei2019miwae} violate the Proposition~\ref{prop-main} (red color indicates the proportion of test samples whose predictive confidence given by the model decreases while providing more modalities, ``CI" is defined in Eq.~\ref{eq:CI}). We estimate the performance on two-modality datasets, and the pie charts show that different samples over-rely on different modalities rather than all samples over-rely on the same modality (e.g., ``$53\%$ Mod1'' indicates ``among the samples who violate  Proposition~\ref{prop-main}, there is $53$ percent of samples whose confidence will increase when Mod2 is removed and the other samples will increase confidence when Mod1 is removed'').}
  \label{fig:observe}
\end{figure*}

\textbf{Multimodal learning} emerges as a promising way to exploit complementary information from different modalities. How to benefit from multimodal data has been a popular research direction, and researchers usually focus on improving architectural designs of the multimodal model~\cite{perez2019mfas,sun2021learning}. In the setting of multimodal classification, MMTM~\cite{joze2020mmtm} achieves state-of-the-art performance by connecting corresponding convolutional layers from different uni-modal branches. Considering the proposed method calibrating confidence with using different number of modalities, multimodal classifiers that can deal with incomplete data are natural candidates to validate our motivation. There is a wide range of research interests in handling missing modalities for multimodal learning, including imputation-independent methods~\cite{zhang2019cpm} and imputation-dependent methods~\cite{mattei2019miwae,wu2018multimodal}. For imputation-independent methods, there is no need to reconstruct the missing modalities and conduct classification using the imputed data. Imputation-dependent methods usually conduct classification with two stages, reconstructing the missing modalities and making classification according to the reconstructed modalities. In this paper, we employ CPM-Nets~\cite{zhang2019cpm}, MIWAE~\cite{mattei2019miwae}, and MMTM~\cite{joze2020mmtm} to validate our motivation due to their representativeness in multimodal learning.

\section{Method}
In this section, we first introduce some basic notations in Section~\ref{sec:preli}. We show the basic assumption of our method and its empirical motivation in Section~\ref{sec:assumption} based on the
principle ``the essence of information is to eliminate uncertainty'', and then evaluate the confidence estimation performance of current multimodal methods in Section~\ref{sec:perform} and find they violate the principle. At the end, we propose a simple yet effective regularization technique to improve the confidence estimation of multimodal models and elaborate the technical details in Section~\ref{sec:regular}.

\subsection{Notation}\label{sec:preli}
We define the training data as $\mathcal{D}=\left\{\{{x}_i^m\}_{m=1}^M,y_i\right\}_{i=1}^N$, where $x_i^m$ is the $m$-th modality of the $i$-th sample, and $y_i \in \{1, \cdots, K\}$ is the corresponding class label. To distinguish one modality or a set of modalities, we use $x^m$ and ${x}^{({\mathbb{S}})}$ to represent the $m$-th modality and multiple modalities respectively, where ${\mathbb{S}}$ is a set of modalities' indexes (e.g., if we have ${\mathbb{S}}=\{ 1, 2 \}$, then ${x}^{({\mathbb{S}})}$ indicates a feature set consisting of $x^1$ and $x^2$, and ${x}^{(\mathbb{M})}=\{ {x}^1, \cdots, {x}^M \}$ indicates the complete $M$ modalities).
The goal is to learn a function parameterized by $\theta$: $f({x}^{(\mathbb{M})},\theta) \rightarrow z$,
where the output $z$ of the network is a vector of $K$ values called logits. Then the logits vector is transformed by a softmax layer: $\hat{p_k} = e^{z_k}/{\sum_k e^{z_k}}$, where the probability distribution of a sample $x$ is defined as ${\mathrm{P}}(y \mid \theta, {x}^{(\mathbb{M})})=\{\hat{p_k}\}_1^K$. The predicted class label is $\hat{y}=\arg \max _{y} {\mathrm{P}}(y \mid \theta, {x}^{(\mathbb{M})})$, and the confidence is defined as $\text{Conf}({x}^{(\mathbb{M})})=\max _{y} {\mathrm{P}}(y \mid \theta, {x}^{(\mathbb{M})})$.

\subsection{Basic Assumption}\label{sec:assumption}
In real-world applications, the quality of multimodal data is usually unstable (e.g., some modalities may be corrupted), so the quality of the multimodal input should be reflected in some quantitative manner (i.e., predictive confidence) which is especially important when multimodal models are deployed for the high-stake tasks. However, it is difficult to exactly define the ``quality'' of each sample, and we can not define the exact functional relationship between the quality and confidence since the confidence from different models is basically different for a same sample. This issue results in the lack of supervision for confidence estimation. Fortunately, according to the principle ``\emph{the essence of information is to eliminate uncertainty (Shannon)}'' in informatics~\cite{soni2017mind,burgin2002essence} (i.e., more information, less uncertainty), we can approximate this relationship with a ranking-based form as follow:

\begin{tcolorbox}[colback=gray!10,
                  colframe=black,
                  width=\linewidth,
                  arc=1mm, auto outer arc,
                  boxrule=0.5pt,
                 ]
\begin{proposition}
 Given two versions of a sample ${x}^{(\mathbb{M})}$, i.e.,  ${x}^{({\mathbb{T}})}$ and ${x}^{({\mathbb{S}})}$, if we can assure ${\mathbb{T}} \subset{\mathbb{S}} \subseteq \mathbb{M}$, then, for a trustworthy multimodal classifier $f(\cdot)$, it should hold $\text{\emph{Conf}}(f({x}^{({\mathbb{T}})})) \leq \text{\emph{Conf}} (f({x}^{({\mathbb{S}})})$.
 \label{prop-main}
\end{proposition}
\end{tcolorbox}

For most multimodal learning tasks, all modalities are assumed to be predictive for the target~\cite{wu2022Characterizing}, and the proposed method is also based on this assumption. For a trustworthy classifier, the predictive confidence should not increase when one modality is removed. We further define the prediction \textbf{C}onfidence \textbf{I}ncrement (CI) with informativeness increment for a sample as:
\begin{equation}
\begin{aligned}
    \mathrm{CI}({x}^{({\mathbb{T}})},{x}^{({\mathbb{S}})}) = \mathrm{Conf}  (f({x}^{({\mathbb{S}})})) - \mathrm{Conf}(f({x}^{({\mathbb{T}})}))& \\ \text{s.t. }{\mathbb{T}} \subset {\mathbb{S}} \subseteq \mathbb{M}&,
    \label{eq:CI}
    \end{aligned}
\end{equation}
where ${\mathbb{T}}$ and ${\mathbb{S}}$ are sets of modalities' indexes. Specifically, a negative value indicates a poor confidence estimation performance where the predictive confidence increases when one modality is removed. To quantify the extent that a learned model violates Proposition~\ref{prop-main}, we introduce a novel measure: \textbf{V}iolating \textbf{R}anking \textbf{R}ate (VRR) as the proportion of test samples whose predictive confidence will increase when removing one modality:

\begin{equation}
\begin{aligned}
    \mathrm{VRR} = \mathbb{E}_{({\mathbb{T},~\mathbb{S}})}\left[\mathbbm{1}\left(\mathrm{CI}({x}^{({\mathbb{T}})},{x}^{({\mathbb{S}})})<0\right)\right] &\\ \text{s.t. }{\mathbb{T}} \subset {\mathbb{S}} \subseteq \mathbb{M}&.
    \label{eq:VRR}
\end{aligned}
\end{equation}
Inspired by prior methods~\cite{moon2020Confidence,toneva2018empirical}, we initialize ${\mathbb{S}}$ as the complete modalities, and obtain ${\mathbb{T}}$ by randomly removing a modality from ${\mathbb{S}}$. Then ${\mathbb{T}}$ is regarded as ${\mathbb{S}}$ for another confidence ranking pair and we repeat this process until there is only one modality remained in ${\mathbb{T}}$ (Please refer to Appendix~\ref{sec:app-pair} for detail). A natural question then arises: how about the confidence estimation performance of the current methods when one modality is removed?

\subsection{Confidence Estimation Performance of Current Multimodal Methods}\label{sec:perform}

To evaluate the quality of confidence estimation of existing multimodal classifiers, we compute the VRR score of CPM-Nets~\cite{zhang2019cpm} and MIWAE~\cite{mattei2019miwae}, which are two typical methods in handling incomplete multimodal data. In addition to classifiers for incomplete multimodal data, we also evaluate MMTM~\cite{wu2022Characterizing}, which is a state-of-the-art multimodal classification method. As shown in Table~\ref{tab:disorder}, the VRR scores of previous methods are quite high which indicates the prediction confidence on a large portion of samples will violate Proposition~\ref{prop-main}. The visualization is shown in Fig.~\ref{fig:observe}, where the red color indicates the proportion of test samples whose predictive confidence estimated by the model decreases while providing more modalities.

A naive strategy is to re-balance the contribution of every modality (i.e., allocating a smaller weight to the modality that samples are over-confident on during the fusion). As shown in Fig.~\ref{fig:observe}, however, we find that different samples are over-confident on different modalities rather than all samples are over-confident on the same modality. This indicates that the problem can not be solved by re-weighting the overall contribution of different modalities since it will make the confidence estimation of some samples worse. Instead, our method characterizes the relationship between the modalities in sample-wise manner, which inherently calibrates the contribution for all samples. Intuitively, it is risky for a model which usually increases the prediction confidence when one modality is removed, since this usually implies that the confidence of the sample and its informativeness are not matched. For this issue, these models can not be deployed into risk-sensitive applications such as medical diagnosis. As a comparison, our method can significantly decrease VRR score (see more details in Table~\ref{tab:disorder}) implying a more trustworthy confidence estimation.

\subsection{Calibrating Multimodal Classification Model}\label{sec:regular}
As shown in Section~\ref{sec:perform}, current multimodal methods usually increase the prediction confidence when one modality is removed, which potentially harms both trustworthiness and performance. To address this issue, the direct strategy is to minimize the following confidence difference:
\begin{equation}
\label{eq:Ldiff}
\mathcal{L}^{({\mathbb{T}},~{\mathbb{S}})} = \mathrm{Conf}({x}^{({\mathbb{T}})})-\mathrm{Conf}({x}^{({\mathbb{S}})}).
\end{equation}
However, models sometimes can still make an accurate prediction confidently when one modality is removed in practice. Eq.~\ref{eq:Ldiff} forces models to produce relatively small confidence when one modality is removed, which results in extremely small confidence for each modality (Please refer to Appendix~\ref{sec:app-diff} for detail). For this issue, we relax this regularization by only penalizing the situation that the estimated confidence increases when one modality is removed. For any pair of multimodal inputs which satisfies that ${\mathbb{T}} \subset {\mathbb{S}} \subseteq \mathbb{M}$, the regularization can be written as:
\begin{equation}
\label{eq:Lq}
\mathcal{L}^{({\mathbb{T}},~{\mathbb{S}})} = \max \left(0, \mathrm{Conf}({x}^{({\mathbb{T}})})-\mathrm{Conf}({x}^{({\mathbb{S}})})\right).
\end{equation}
For each sample, the total regularization loss is integrated over all pairs of inputs with different numbers of modalities, which is formalized as:
\begin{equation}
    \mathcal{L}^\text{CML} = \sum_{({\mathbb{T}},~{\mathbb{S}})} \mathcal{L}^{({\mathbb{T}},~{\mathbb{S}})}, \quad \{ \forall ({\mathbb{T}},~{\mathbb{S}})| {\mathbb{T}} \subset {\mathbb{S}} \subseteq \mathbb{M}\}.
\end{equation}
The exact computation of above loss needs to enumerate all modality set pairs $({\mathbb{T}}$, ${\mathbb{S}})$, which is typically computational expensive sometimes. Therefore, we propose to approximate this loss by sampling and it works well in practice. Specifically, we conduct sampling as same as that in computing VRR defined in Eq.~\ref{eq:VRR}. 

The proposed regularization is general and thus can be equipped by current multimodal classifiers to calibrate their confidence estimation as an additional loss item. We typically provide examples in utilizing the proposed technique in imputation-independent method (i.e., CPM-Nets~\cite{zhang2019cpm}), imputation-dependent method (i.e., MIWAE~\cite{mattei2019miwae}), and recent multimodal classification method (i.e., MMTM~\cite{wu2022Characterizing}). The proposed regularization can be deployed to current multimodal methods flexibly, and accordingly the objective function is induced as:
\begin{equation}
    \mathcal{L} = \mathcal{L}^\text{CL} +  \lambda \mathcal{L}^\text{CML},
\end{equation}
where $\mathcal{L}^\text{CL}$ is the classification loss criterion (e.g., cross-entropy loss), and $\lambda$ is hyperparameter controlling the strength of CML regularization. The process of calibrating multimodal classification are shown in Algorithm~\ref{alg:CML}. 
\begin{algorithm}[ht]
 	\caption{Calibrating Multimodal Classifier}
 	 	\label{alg:CML}
    \begin{algorithmic}
 	\STATE	\textbf{Given} dataset $\mathcal{D}=\left\{\{{x}_i^m\}_{m=1}^M,y_i\right\}_{i=1}^N$, initialized classifier $f$, classification loss criterion $\mathcal{L}^\text{CL}$, hyperparameter $\lambda$, and epochs for training the classifier ${train\_epochs}$.
 		
   \FOR{$e=1,\ldots,{train\_epochs}$}{\STATE ${\mathbb{S}}\leftarrow \mathbb{M}$;~
 		$\mathcal{L}^\text{CL} \leftarrow \mathcal{L}^\text{CL}({x}^{(\mathbb{S})})$;~
 		$\mathcal{L}^\text{CML} \leftarrow 0$;
 		
  \FOR{$m=1,\ldots,M-1$}{
  \STATE Randomly remove a modality of ${\mathbb{S}}$ and set it as ${\mathbb{T}}$;

 	\STATE	Compute the classification loss: 
 		$\mathcal{L}^\text{CL} \leftarrow \mathcal{L}^\text{CL} + \mathcal{L}^\text{CL}({x}^{({\mathbb{T}})})$;

 	\STATE	Compute the regularization loss:	\textcolor{violet}{$\mathcal{L}^\text{CML} \leftarrow \mathcal{L}^\text{CML} + \max\left(0, \mathrm{Conf}({x}^{({\mathbb{T}})})-\mathrm{Conf}({x}^{({\mathbb{S}})}\right)$};
 		
 		$\mathbb{S} \leftarrow \mathbb{T}$;
 		}
   \ENDFOR
 		
 	\STATE	Total loss: $\mathcal{L}=\frac{1}{M}\mathcal{L}^\text{CL}\textcolor{violet}{ +\lambda \mathcal{L}^\text{CML}}$;
 		
 	\STATE	Update the parameters of the classifier $f$ with $\mathcal{L}$;
 		}		
   \ENDFOR
\STATE \textbf{return} the classifier $f$
\end{algorithmic}
\end{algorithm}

\subsection{Discussion and Analyses}

$\circ$ \textbf{Why should a model meet the ranking relationship regardless of class labels?} For multimodal learning, all modalities are assumed to be predictive of the target~\cite{wu2022Characterizing}, which can be expressed as $I(y,{x}^{m}) \geq 0$, where $I(\cdot)$ denotes mutual information~\cite{blum1998combining} and ${x}^{m}$ indicates the $m$-th modality.
\begin{lemma}
    Suppose we have two versions of a sample ${x}^{(\mathbb{M})}$, i.e.,  ${x}^{({\mathbb{T}})}$ and ${x}^{({\mathbb{S}})}$, if we can assure ${\mathbb{T}} \subset{\mathbb{S}} \subseteq \mathbb{M}$, then, for any class label ${y}$, we have $I({y},{x}^{({\mathbb{T}})}) \leq I({y},{x}^{({\mathbb{S}})})$.
\end{lemma}
In other words, ${x}^{({\mathbb{S}})}$ is more predictive for the target than ${x}^{({\mathbb{T}})}$ regardless of the label. For a trustworthy multimodal classification model, the confidence of ${x}^{({\mathbb{T}})}$ should not be larger than ${x}^{({\mathbb{S}})}$.

$\circ$ \textbf{Why can CML regularization calibrate a model?} CML regularization can guarantee a smaller confidence of ${x}^{({\mathbb{T}})}$ when the model makes a wrong prediction of ${x}^{({\mathbb{S}})}$, which means that CML can alleviate the over-confidence.
\begin{lemma}\label{lem:3-2}
    Suppose the CML regularization can achieve a lower $\mathrm{VRR}$, i.e., $\mathrm{VRR}_{CML} < \mathrm{VRR}_\text{ORIG}$, then for the samples that meet $\mathbb{E}\left(\mathrm{Conf}_{CML}({x}^{(\mathbb{S})})\right)=\mathbb{E}\left(\mathrm{Conf}_\text{ORIG}({x}^{(\mathbb{S})})\right)$, we have $\mathbb{E}\left(\mathrm{Conf}_{CML}({x}^{(\mathbb{T})})\right)\leq\mathbb{E}\left(\mathrm{Conf}_\text{ORIG}({x}^{(\mathbb{T})})\right)$.
\end{lemma} 
From the empirical results, we find $\mathrm{Conf}_{CML}({x}^{(\mathbb{S})})$ and $\mathrm{Conf}_{ORIG}({x}^{(\mathbb{S})})$ are very similar for most samples, where $\mathrm{Conf}_{ORIG}(\cdot)$ and $\mathrm{Conf}_{CML}(\cdot)$ indicate the confidence estimated by the original (ORIG) model and the model improved by CML regularization respectively. The proof of Lemma~\ref{lem:3-2} and empirical results please refer to Appendix~\ref{sec:app-conf}.

$\circ$ \textbf{Why not just penalize the difference in confidence (i.e., minimizing $\text{Conf}({x}^{({\mathbb{T}})})-\text{Conf}({x}^{({\mathbb{S}})})$)?} Forcing the confidence for ${x}^{({\mathbb{T}})}$ to be smaller than the confidence for ${x}^{({\mathbb{S}})}$ regardless of whether the samples violate the Prop.~\ref{prop-main} will lead to very small confidence for ${x}^{({\mathbb{T}})}$, and adding such a penalty to samples who meet the Prop.~\ref{prop-main} will lead to a trivial solution (i.e., extremely small confidence when any modality is removed, and the experiments are shown in Appendix~\ref{sec:app-diff}). What's more, the model sometimes can still make correct predictions confidently when one modality is removed. A flexible ranking regularization (Eq.~\ref{eq:Lq}) makes it more reasonable for the real situation.

\begin{table*}[th]

\begin{center}
\caption{VRR ($\%$) of test samples (a lower value indicates a better confidence estimation. Type III is shown in Appendix). ``\xmark'' indicates the model is not equipped with the proposed regularization ($\lambda=0$). Performance on Type III please refer to Appendix~\ref{sec:app-diff}.}
\label{tab:disorder}
\center
\resizebox{1.0\textwidth}{!}{
 \setlength{\tabcolsep}{3.9mm}
 \begin{tabular}{c|ccccccc}
\toprule
\multicolumn{1}{c}{\text{Method}}   & \text{CML}   & \text{TUANDROMD} & \text{YaleB}& \text{Handwritten} & \text{CUB} &\text{Animal} \\ \midrule
\multirow{3}{*} {$\text{Type I}$}   & \xmark    &  $23.38\pm1.39$                                                    &$39.15\pm4.97$  &$17.64\pm2.31$   &$2.83\pm1.55$   &$44.39\pm7.55$                                                     \\
                                                                                  
        & \cmark                                                                                   &$12.58\pm2.84$                                                    &$15.05\pm1.12$  &$~~3.18\pm0.80$   &${2.17\pm1.13}$   &${29.02\pm5.43}$     \\ & Improve & \textcolor{mycolor2}{{$~~~~~~\bigtriangleup 10.80 $}} & \textcolor{mycolor2}{{$~~~~~~\bigtriangleup 24.10 $}}& \textcolor{mycolor2}{{$~~~~~~\bigtriangleup 14.46 $}}& \textcolor{mycolor2}{{$~~~~~~~~\bigtriangleup 0.66 $}}& \textcolor{mycolor2}{{$~~~~~~\bigtriangleup 15.37 $}}                               \\ \midrule \multirow{3}{*}{{$\text{Type II}$}}
     & \xmark       &$39.17\pm2.32$ &$20.54\pm4.26$    &$33.82\pm5.16$     &$23.17\pm4.87$    &$12.51\pm1.50$                    \\
     & \cmark                                                    &$~~{8.38\pm1.31}$   &${14.46\pm2.17}$  &${29.99\pm2.30}$  &${20.17\pm3.05}$  &${  ~~8.64\pm0.32}$    \\& Improve & \textcolor{mycolor2}{{$~~~~~~\bigtriangleup 30.79 $}} & \textcolor{mycolor2}{{$~~~~~~~~\bigtriangleup 6.08 $}}& \textcolor{mycolor2}{{$~~~~~~~~\bigtriangleup 3.83 $}}& \textcolor{mycolor2}{{$~~~~~~~~\bigtriangleup 3.00 $}}& \textcolor{mycolor2}{{$~~~~~~~~\bigtriangleup 3.87 $}}                             \\\bottomrule
\end{tabular}}
\end{center}

\end{table*}

\section{Experiments}

\subsection{Setup}
\begin{table*}[ht]
\vskip 0.15in
\begin{center}
\small
\caption{Accuracy performance comparison for whether the model is equipped with the CML regularization term (i.e., whether $\lambda$ is set to 0). The means and standard deviations over five runs are reported.}
\label{tab:acc}
\center
\resizebox{1.0\textwidth}{!}{
\setlength{\tabcolsep}{3.9mm}
\begin{tabular}{c|c|ccccc}
\toprule
\multicolumn{1}{c}{\text{Method}}  & \multicolumn{1}{c}{\text{Dataset}}  & \text{CML}  & \text{\begin{tabular}[c]{@{}c@{}}Accuracy\\ ($\uparrow$)\end{tabular}} & \text{\begin{tabular}[c]{@{}c@{}}NLL\\ ($\downarrow$)\end{tabular}} & \text{\begin{tabular}[c]{@{}c@{}}AURC\\ ($\downarrow$)\end{tabular}}& \text{\begin{tabular}[c]{@{}c@{}}E-AURC\\ ($\downarrow$)\end{tabular}} \\ \midrule 
\multirow{10}{*} {\text{Type I}}         
     &  \multirow{3}{*}{\text{CUB}}      & \xmark          &$87.00\pm4.36$                                                    &$20.49\pm0.30$  &$59.44\pm22.10$   &$49.52\pm17.35$                         \\
                                                                                 &             & \cmark                                                    &${88.33\pm4.05}$                                                    &${20.53\pm0.46}$  &${55.94\pm17.07}$   &${47.92\pm16.89}$       \\& & Improve & \textcolor{mycolor4}{{$~~~~~~~~\bigtriangleup 1.33 $}} & \textcolor{mycolor1}{{$~~~~~~~~\bigtriangledown 0.04 $}}& \textcolor{mycolor4}{{$~~~~~~\bigtriangleup 3.50 $}}& \textcolor{mycolor4}{{$~~~~~~~\bigtriangleup 1.60 $}}                                          \\ \cmidrule{2-7}
    
     &  \multirow{3}{*}{\text{Animal}}      & \xmark          &$81.72\pm2.51$                                                    &$36.87\pm0.41$  &$82.14\pm27.20$   &$63.94\pm22.74$                              \\
                                                                                 &             & \cmark                                                    &${82.73\pm1.64}$                                                    &$36.87\pm0.36$  &${71.54\pm16.03}$   &${55.50\pm13.13}$   \\& & Improve & \textcolor{mycolor4}{{$~~~~~~~~\bigtriangleup 1.01 $}} & \textcolor{mycolor4}{$~~~~~~~~~~~0.00$}& \textcolor{mycolor4}{{$~~~~~~\bigtriangleup 10.60 $}}& \textcolor{mycolor4}{{$~~~~~~~\bigtriangleup 8.44 $}}     \\ \cmidrule{2-7}
     &  \multirow{3}{*}{\shortstack{\text{TUAND-}\\\text{ROMD}}}      & \xmark          &$84.66\pm0.43$                                                    &$6.88\pm0.00$  &$61.46\pm6.09$   &$49.00\pm5.75$                                            \\
                                                                                 &             & \cmark                                                    &${85.20\pm0.81}$                                                   &${6.88\pm0.00}$  &${58.24\pm5.05}$   &${46.64\pm4.55}$   \\& & Improve & \textcolor{mycolor2}{{$~~~~~~~~\bigtriangleup 0.54 $}} & \textcolor{mycolor4}{$~~~~~~~~~~~0.00$}& \textcolor{mycolor4}{{$~~~~~~\bigtriangleup 3.22 $}}& \textcolor{mycolor4}{{$~~~~~~~\bigtriangleup 2.36 $}}                                                       \\ \midrule
\multirow{10}{*} {\text{Type II}}    
     
     &  \multirow{3}{*}{\text{CUB}}      & \xmark          &$92.33\pm1.11$                                                    &$2.33\pm0.55$  &$10.92\pm1.94$   &$7.82\pm1.32$                    \\
                                                                                 &             & \cmark                                                    &${94.50\pm1.71}$                                                    &${2.24\pm1.27}$  &${9.32\pm3.91}$   &${7.60\pm3.02}$        \\& & Improve & \textcolor{mycolor2}{{$~~~~~~~~\bigtriangleup 2.17 $}} & \textcolor{mycolor4}{{$~~~~~~\bigtriangleup 0.09 $}}& \textcolor{mycolor2}{{$~~~~~~\bigtriangleup 1.60 $}}& \textcolor{mycolor4}{{$~~~~~~~\bigtriangleup 0.22 $}}                                        \\ \cmidrule{2-7}
    
     &  \multirow{3}{*}{\text{Animal}}      & \xmark          &$86.75\pm0.33$                                                  &$8.25\pm3.79$  &$27.62\pm7.42$   &$18.40\pm7.27$               \\
                                                                                 &             & \cmark                                                    &${87.61\pm0.50}$                                                    &${4.99\pm0.46}$  &${21.26\pm1.31}$   &${13.24\pm0.92}$     \\& & Improve & \textcolor{mycolor2}{{$~~~~~~~~\bigtriangleup 0.86 $}} & \textcolor{mycolor2}{{$~~~~~~\bigtriangleup 3.26 $}}& \textcolor{mycolor2}{{$~~~~~~\bigtriangleup 6.36 $}}& \textcolor{mycolor2}{{$~~~~~~~\bigtriangleup 5.16 $}} \\ \cmidrule{2-7}
     &  \multirow{3}{*}{\shortstack{\text{TUAND-}\\\text{ROMD}}}      & \xmark          &$86.32\pm0.85$                                                    &$3.26\pm0.09$  &$43.40\pm2.65$   &$33.56\pm2.38$               \\
                                                                                 &             & \cmark                                                    &${88.69\pm0.99}$                                                    &${3.21\pm0.15}$  &${38.62\pm5.44}$   &${31.90\pm4.37}$   \\& & Improve & \textcolor{mycolor2}{{$~~~~~~~~\bigtriangleup 2.37 $}} & \textcolor{mycolor4}{{$~~~~~~\bigtriangleup 0.02 $}}& \textcolor{mycolor2}{{$~~~~~~\bigtriangleup 4.78 $}}& \textcolor{mycolor4}{{$~~~~~~~\bigtriangleup 1.66 $}}                                                        \\
       \midrule
\multirow{7}{*} {\text{Type III}}    
     
     &  \multirow{3}{*}{\text{NYUD2}}      & \xmark          &$66.89\pm0.85$                                                    &$10.03\pm0.10$  &$140.53\pm5.66$   &$78.40\pm5.01$                    \\
                                                                                 &             & \cmark                                                    &${68.09\pm0.68}$                                                    &${9.83\pm0.15}$  &${137.27\pm6.94}$   &${79.87\pm6.30}$        \\& & Improve & \textcolor{mycolor2}{{$~~~~~~~~\bigtriangleup 1.20 $}} & \textcolor{mycolor2}{{$~~~~~~\bigtriangleup 0.20 $}}& \textcolor{mycolor4}{{$~~~~~~\bigtriangleup 3.26 $}}& \textcolor{mycolor1}{{$~~~~~~~~\bigtriangledown 1.47 $}}
                                                                                 
                                                                                 \\
                                                                                 
                                                                        \cmidrule{2-7}
     &  \multirow{3}{*}{\shortstack{\text{SUN-}\\\text{RGBD}}}      & \xmark          &$62.11\pm0.31$                                                    &$13.27\pm0.53$  &$181.00\pm1.20$   &$97.87\pm1.48$               \\
                                                                                 &             & \cmark                                                    &${62.78\pm0.32}$                                                    &${13.25\pm0.46}$  &${174.90\pm1.50}$   &${95.00\pm1.00}$   \\& & Improve & \textcolor{mycolor2}{{$~~~~~~~~\bigtriangleup 0.67 $}} & \textcolor{mycolor4}{{$~~~~~~\bigtriangleup 0.05 $}}& \textcolor{mycolor2}{{$~~~~~~\bigtriangleup 6.10 $}}& \textcolor{mycolor2}{{$~~~~~~~\bigtriangleup 2.87 $}}                                                        \\
                                                                                 \bottomrule
\end{tabular}}
\end{center}
\end{table*}

We deploy the proposed regularization strategy into different types of multimodal classifiers including the imputation-independent method (Type I), the imputation-dependent method (Type II), and the recent state-of-the-art method (Type III). CPM-Nets~\cite{zhang2019cpm} is a typical imputation-independent algorithm, which can adapt to arbitrary missing patterns without reconstructing the missing modalities. MIWAE~\cite{mattei2019miwae} is a imputation-dependent algorithm. The above two methods are well-established models in incomplete multimodal learning. In addition to incomplete multimodal learning methods, we also deploy the regularization into an advanced multimodal classification method~\cite{wu2022Characterizing}, which is termed Multimodal Transfer Module (MMTM). We approximate the modality removal by feature corruption (e.g., adding strong noise) because MMTM can not make a prediction when one modality is explicitly removed.
For a fair comparison, the only difference between whether the model is equipped with CML regularization or not. Please refer to Appendix~\ref{sec:app-setting} for more detailed settings.

\textbf{Datasets:}We evaluate the proposed method on diverse datasets, including data with multimodal data, such as YaleB~\cite{geo2002From}, Handwritten~\cite{perkins2003Online}, CUB~\cite{wah2011caltech}, Animal~\cite{krizhevsky2012imagenet,simonyan2014very} (which is a dataset under class-imbalanced), TUANDROMD~\cite{borah2020malware}, NYUD2~\cite{qi20173d}, and SUNRGBD~\cite{song2015sun}. It should be pointed out that we also estimate the proposed method on the class-imbalanced dataset. We find that CML can improve the performance when the training data is class-imbalanced since CML calibrates the model regardless of the label while the vanilla model always tends to be under-confidence of the minority classes compared with majority classes. For more detailed analysis please refer to Appendix~\ref{sec:app-datasets}.


\subsection{Questions to be Verified}
We conduct diverse experiments to comprehensively investigate the underlying assumption and the proposed method, including:

$\circ$ \textbf{Can CML regularization improve the confidence estimation of multimodal classifiers?} To validate whether the proposed method improves multimodal classifiers' confidence estimation, we evaluate the confidence estimation of current multimodal classifiers without and with CML regularization, respectively. We conduct experiments of each type of method on seven datasets and evaluate their trustworthiness in terms of VRR (defined in Eq.~\ref{eq:VRR}).

$\circ$ \textbf{Can CML regularization improve robustness?} CML regularization can improve multimodal classifiers' confidence estimation, so a natural question arises - does a better confidence estimation imply better robustness? To verify this, we evaluate the robustness on the complete multimodal data and noisy multimodal data (adding Gaussian noise to some modalities, i.e., zero mean with varying variance $\epsilon$).

$\circ$ \textbf{Is CML easy to be deployed and not sensitive to hyperparameters?} In order to investigate the key factor that makes the improvement in the proposed method, we evaluate the performance in terms of classification accuracy under different strengths of CML regularization. We conduct experiments on both the original and noised data (i.e., adding noise to one of the modalities during the test). More details are shown in Appendix~\ref{sec:app-setting}.

\subsection{Results}
\subsubsection{Confidence Estimation}
\begin{figure}[ht]
\centering
  \subfloat[Animal]{
  \centering
  \includegraphics[width=0.47\linewidth]{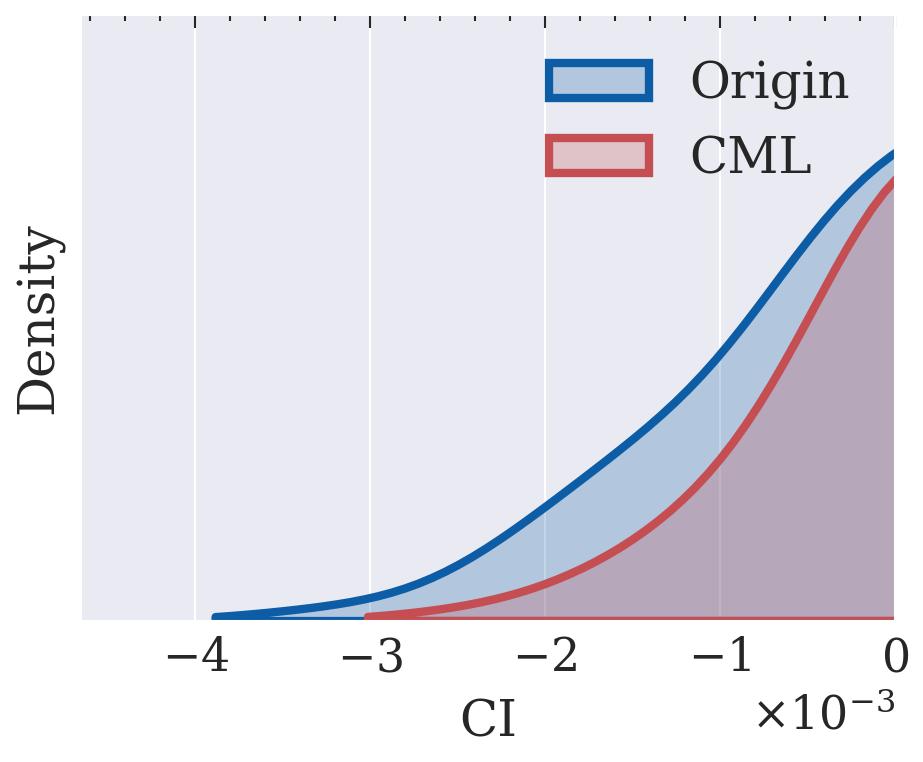}
  }
  \centering
  \subfloat[Tuandromd]{
  \centering
  \includegraphics[width=0.47\linewidth]{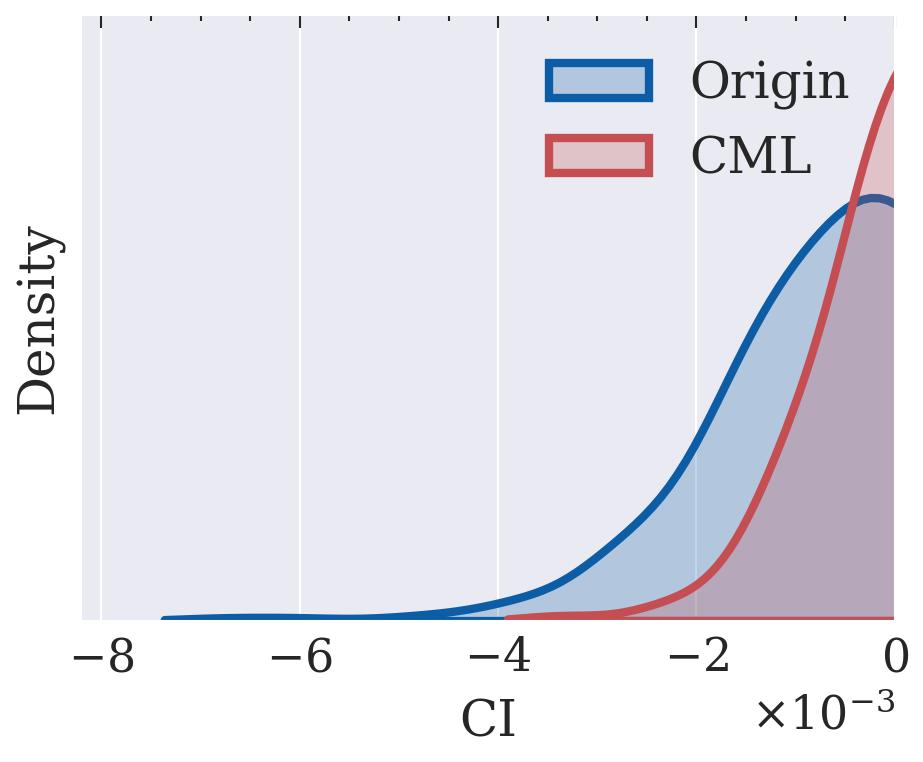}
  }
 \caption{Confidence estimation when one modality is removed, where ``CI'' is defined in Eq.~\ref{eq:CI}.}
 \label{fig:diff}
\end{figure}

\begin{table*}[!th]

\begin{center}
\caption{Accuracy performance comparison when some of the modalities is corrupted with Gaussian noise (i.e., zero mean with varying variance $\epsilon$).}
\label{tab:noise}
\center
\resizebox{1.0\textwidth}{!}{
\setlength{\tabcolsep}{3.9mm}
\begin{tabular}{c|c|ccccc}
\toprule
\multicolumn{1}{c}{\text{Dataset}}  & \multicolumn{1}{c}{\text{Noise on}}  & \multicolumn{1}{c}{\text{CML}}  &
\text{$\mathbf{\epsilon=0.1}$}& \text{\begin{tabular}[c]{@{}c@{}}$\mathbf{\epsilon=0.2}$\\ \end{tabular}}& \text{\begin{tabular}[c]{@{}c@{}}$\mathbf{\epsilon=0.3}$\\ \end{tabular}}& \text{\begin{tabular}[c]{@{}c@{}}$\mathbf{\epsilon=0.5}$\\ \end{tabular}}
\\ \midrule
\multirow{10}{*} {\text{CUB}}  
    &   \multirow{3}{*}{\text{\{1\}}}      & \xmark                                         &$84.72\pm3.32$  
       &$82.22\pm4.53$   
       &$79.72\pm4.43$
       &$71.17\pm9.14$ \\
                                                                                 &           & \cmark            
                                                         &${85.83\pm2.72}$   
                                                         &${85.00\pm3.50}$   
                                                         &${84.17\pm4.08}$ 
                                                         &${81.11\pm4.37}$               \\& & Improve & \textcolor{mycolor4}{{$~~~~~~~~\bigtriangleup 1.11 $}} & \textcolor{mycolor4}{{$~~~~~~\bigtriangleup 2.78 $}}& \textcolor{mycolor2}{{$~~~~~~\bigtriangleup 4.45 $}}  & \textcolor{mycolor2}{{$~~~~~~\bigtriangleup 9.94 $}}                     \\ \cmidrule{2-7}
      &  \multirow{3}{*}{\text{\{2\}}}       & \xmark                                                    &$84.44\pm2.75$  &$83.89\pm3.22$   &$83.61\pm2.83$   &$83.61\pm3.87$    \\
                                                                                  
                                                            &                           & \cmark                                                                                            &${85.83\pm3.40}$   &${85.28\pm2.75}$   &${85.28\pm1.97}$  &${85.00\pm1.80}$   \\& & Improve & \textcolor{mycolor4}{{$~~~~~~~~\bigtriangleup 1.39 $}} & \textcolor{mycolor4}{{$~~~~~~\bigtriangleup 1.39 $}}& \textcolor{mycolor4}{{$~~~~~~\bigtriangleup 1.67 $}}  & \textcolor{mycolor4}{{$~~~~~~\bigtriangleup 1.39 $}}                                     \\ \cmidrule{2-7}
     &  \multirow{3}{*}{\text{\{1, 2\}}}      & \xmark                                                      &$85.00\pm3.12$  &$82.78\pm3.98$  &$80.00\pm4.46$  &$72.50\pm11.14$                                         \\
                                                                                 &             & \cmark                                                       &${85.83\pm2.72}$  &${85.84\pm3.12}$   &${85.83\pm4.25}$         &${81.39\pm6.43~~}$                        \\& & Improve & \textcolor{mycolor4}{{$~~~~~~~~\bigtriangleup 0.83 $}} & \textcolor{mycolor4}{{$~~~~~~\bigtriangleup 3.06 $}}& \textcolor{mycolor2}{{$~~~~~~\bigtriangleup 5.83 $}}  &  \textcolor{mycolor2}{{$~~~~~~\bigtriangleup 8.89 $}}                                                                              \\ \midrule
\multirow{10}{*} {\text{Animal}}  
     &  \multirow{3}{*}{\text{\{1\}}}      & \xmark                                                             &$80.78\pm2.79$   &$80.96\pm2.78$   &$80.85\pm2.80$        &$80.68\pm2.93$                           \\
                                                                                 &             & \cmark                                                 &${82.03\pm1.91}$  &${82.37\pm2.09}$   &${82.55\pm2.24}$      &${82.30\pm2.40}$ \\& & Improve & \textcolor{mycolor4}{{$~~~~~~~~\bigtriangleup 1.25 $}} & \textcolor{mycolor4}{{$~~~~~~\bigtriangleup 1.41 $}}& \textcolor{mycolor4}{{$~~~~~~\bigtriangleup 1.70 $}}  &  \textcolor{mycolor4}{{$~~~~~~\bigtriangleup 1.62 $}}                                      \\ \cmidrule{2-7}
      &  \multirow{3}{*}{\text{\{2\}}}       & \xmark                                                     &$80.70\pm2.45$  &$79.81\pm3.14$   &$77.34\pm4.80$  &     $68.52\pm9.68$    \\
                                                            &                       & \cmark                           &${82.07\pm1.57}$  &${81.23\pm2.32}$   &${78.93\pm3.65}$           &${72.39\pm8.35}$    \\& & Improve & \textcolor{mycolor4}{{$~~~~~~~~\bigtriangleup 1.37 $}} & \textcolor{mycolor4}{{$~~~~~~\bigtriangleup 1.42 $}}& \textcolor{mycolor4}{{$~~~~~~\bigtriangleup 1.59 $}}  &  \textcolor{mycolor4}{{$~~~~~~\bigtriangleup 3.87 $}}                                    \\ \cmidrule{2-7}
     &  \multirow{3}{*}{\text{\{1, 2\}}}      & \xmark                                                        &$80.87\pm2.55$   &$79.97\pm3.12$   &$77.11\pm5.86$        &$65.08\pm12.75$                                 \\
                                                                                 &             & \cmark                                       &${82.14\pm1.76}$
                                          &${81.95\pm2.65}$               &${79.63\pm5.28}$      &${72.46\pm11.39}$                                               \\& & Improve & \textcolor{mycolor4}{{$~~~~~~~~\bigtriangleup 1.27 $}} & \textcolor{mycolor4}{{$~~~~~~\bigtriangleup 1.98 $}}& \textcolor{mycolor4}{{$~~~~~~\bigtriangleup 2.52 $}}  &  \textcolor{mycolor4}{{$~~~~~~\bigtriangleup 7.38 $}}                                         \\   \bottomrule
\end{tabular}}
\end{center}
\end{table*}

We evaluate the confidence estimation of current multimodal learning models from a ranking perspective. It is observed that for a large portion of samples the confidence will increase when one modality is removed, while the confidence estimation of the classification models equipped with our proposed CML regularization is significantly improved. We intuitively demonstrate the confidence changing in Fig.~\ref{fig:diff}, and the quantitative results are shown in Tab.~\ref{tab:disorder}. According to Fig.~\ref{fig:diff}, we show the confidence estimation of CPM-Nets, where ``Original'' and ``CML'' indicate the model is without and with the proposed CML regularization respectively. According to Fig.~\ref{fig:diff}, it is observed that the confidence without CML regularization may increase when one modality is removed, which indicates that the model fails to take all modalities into account fairly when making predictions. This will lead to unpromising robustness and generalization, which clearly verifies the main assumption in Sec.~\ref{sec:robustness}.


\subsubsection{CML Regularization Improves Robustness}\label{sec:robustness}
In this subsection, we evaluate the performance on the complete multimodal data, where the training/test data is divided as previous work~\cite{zhang2019cpm}. From Tab.~\ref{tab:acc}, the classification models equipped with CML regularization consistently outperform their counterparts (i.e., the original classification models) validating the rationality of CML principle. It is worth noting that Type III exhibits a significant improvement, while the improvement in Type I and Type II is relatively minor compared to the standard deviation. The high variance can be attributed to the baseline models themselves. To avoid the influence of empirical contingency, we report the means and standard deviations over 5 or 10 runs in our paper. Furthermore, we distinguish the marks in the table based on the significance of the improvement, with a lighter color indicating a relatively minor improvement compared to the standard deviation. Results on more datasets are shown in Appendix~\ref{sec:app-acc}.

Significantly improving the accuracy on real-world data without additional techniques or more advanced architectures can be challenging as the benchmark datasets have already achieved good performance in terms of accuracy. However, we observed that the models equipped with CML regularization are more robust to noise, particularly when the noise is heavy. Specifically, we find that CML regularization can improve the robustness of imperfect data, such as noise. We evaluate the models in terms of the accuracy in the test under Gaussian noise (i.e., zero mean and varying variance $\epsilon$), and ``Noise On'' indicates which modality is noised (e.g., $\{1\}$ indicates the first modality is noised). We report the performance on the challenging datasets (CUB and Animal) in the main text (Tab.~\ref{tab:noise}) and more results are in Appendix~\ref{sec:app-noise}. We can find that the models equipped with CML regularization are more robust to noise, especially when the noise is much heavier. 

\subsubsection{Performance under Different Strengths of CML Regularization}\label{sec:ablation}

In this subsection, we report the accuracy under different strengths of regularization (where ``$\lambda=0$'' indicates the model is not equipped with the proposed CML regularization). We also add Gaussian noise (i.e., zero mean and varying variance $\epsilon$) to one of the modalities on CUB, and it is clear that the model with CML regularization is more robust to the potential noise.

\begin{figure}[!ht] 
  \centering
  \subfloat[Noise on the first modality]{
  \includegraphics[width=0.23\textwidth]{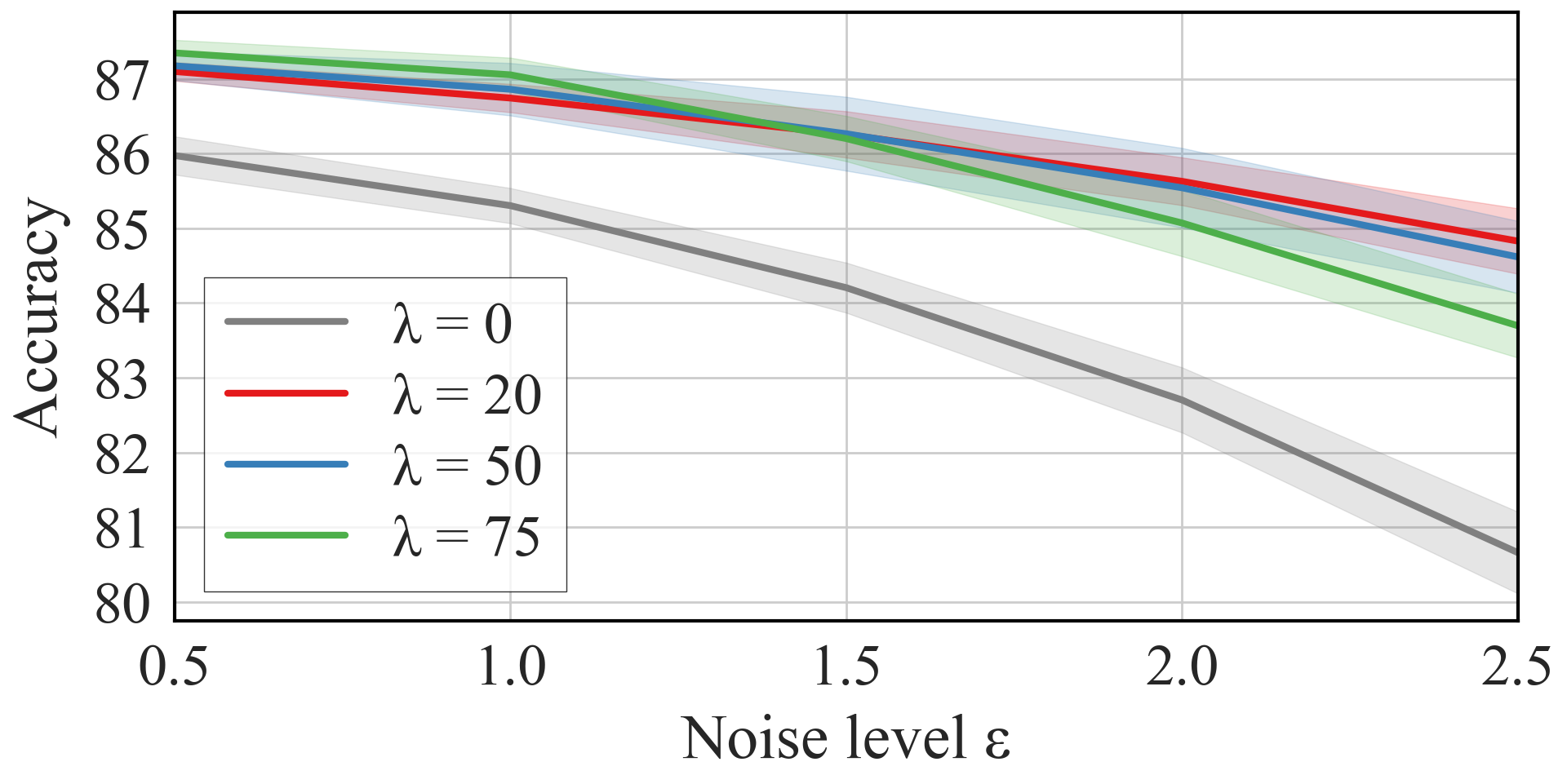}
  }
    \subfloat[Noise on the second modality]{
  \includegraphics[width=0.23\textwidth]{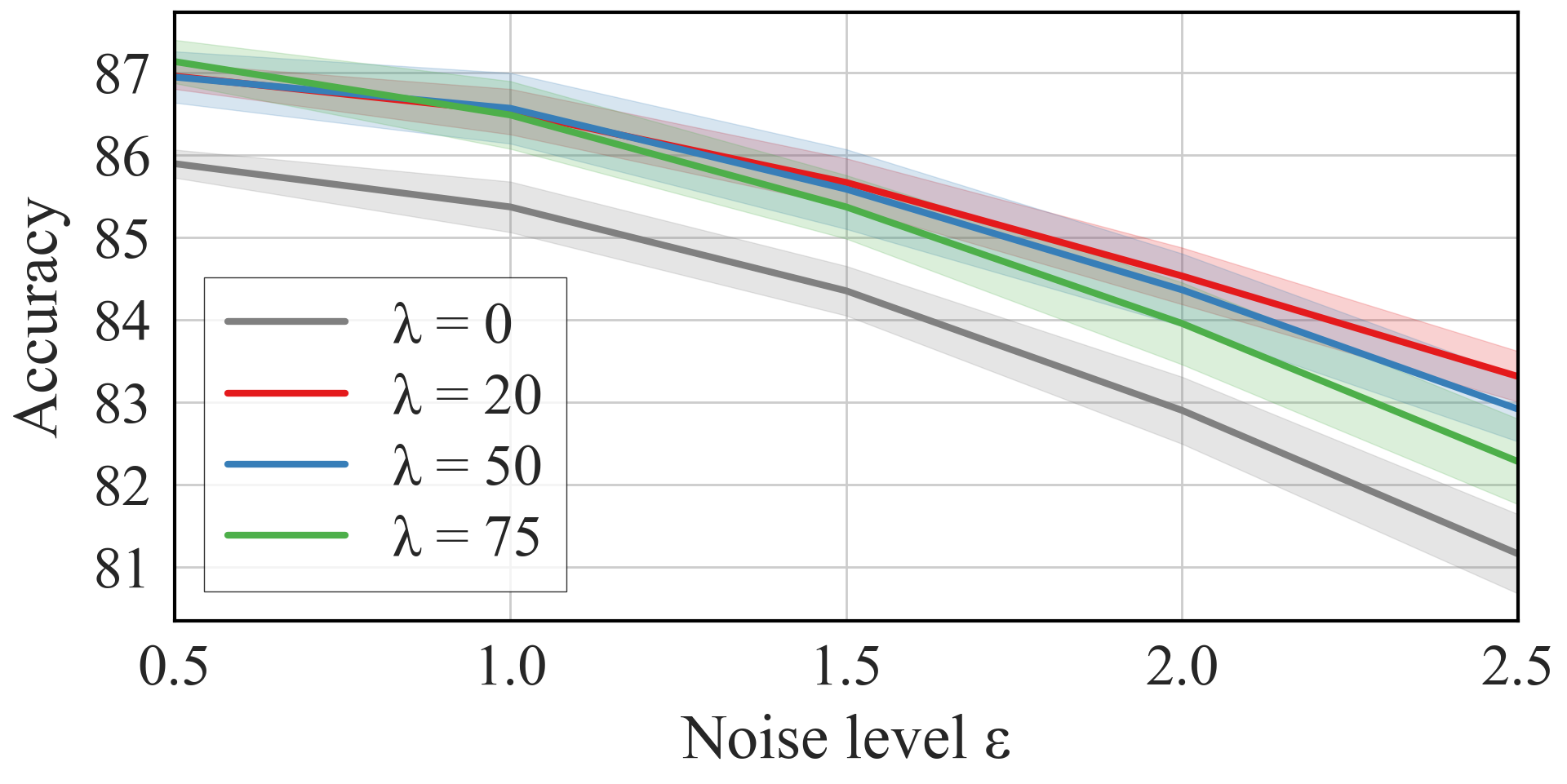}
  }
 \caption{Accuracy estimation where one of the modalities is corrupted with noise.}
  \label{fig:noise}
\end{figure}
As shown in Fig.~\ref{fig:noise}, it is observed that CML regularization can promote accuracy on the noisy data. The potential reason is that the CML regularization enforces the reasonable confidence estimation and thus prohibits the model from being over-confident on the low-quality modality, where the low-quality modality usually tends to result in a wrong decision. Moreover, according to Fig.~\ref{fig:noise}, the proposed regularization is not sensitive to the hyperparameter $\lambda$, where promising performance could be expected with a mild regularization strength. In other words, the proposed regularization is not sensitive to hyperparameters and CML is easy to be deployed into a wide spectrum of multimodal models.

\section{Conclusion}

In this work, we reveal a novel issue widely existing in multimodal learning through extensive empirical studies. We observe that the confidence estimations of current multimodal learning algorithms are typically unreliable, and tend to rely on some partial modalities. This further results in the non-robustness of learned models against modality corruption. Concretely, existing multimodal classifiers tend to be overconfident based on some modalities, and ignore the valuable evidence from other modalities even those might be critical to make the decision. To solve this problem, we introduce a novel regularization technique to calibrate the confidence estimation, which forces model to estimate a calibrated predictive confidence. This technique can be naturally deployed into existing multimodal learning methods without modifying the main training process. We conduct comprehensive experiments which demonstrate the superiority of our method in classification in terms of both accuracy and calibration. The proposed method is the first attempt to calibrate the relationship between confidence and the number of modalities used in multimodal learning. This research is an inspirational topic which could benefit the multimodal learning community. In current implementation, we employ sampling to construct constraint. Although it is widely used and effective in machine learning, we will focus on more principled approximation strategies in the future.
\section*{Acknowledgments}
This work is jointly supported by the National Natural
Science Foundation of China (Grant No. 61976151), the Agency for Science, Technology and Research (A*STAR) under
its AME Programmatic Funding Scheme (Project
No. A18A1b0045), and A*STAR Central Research Fund. We gratefully acknowledge the support of CAAI-Huawei MindSpore Open Fund\footnote{https://www.mindspore.cn/}. The project was finished during the internship in AI Lab, Tencent.

\bibliography{example_paper}
\bibliographystyle{icml2023}

\newpage
\appendix
\onecolumn

\etocdepthtag.toc{mtappendix}
\etocsettagdepth{mtchapter}{none}
\etocsettagdepth{mtappendix}{subsection}


\section{How to Make Ranking Pairs}\label{sec:app-pair}

\begin{figure}[ht]
    \centering
    \includegraphics[width=0.94\textwidth]{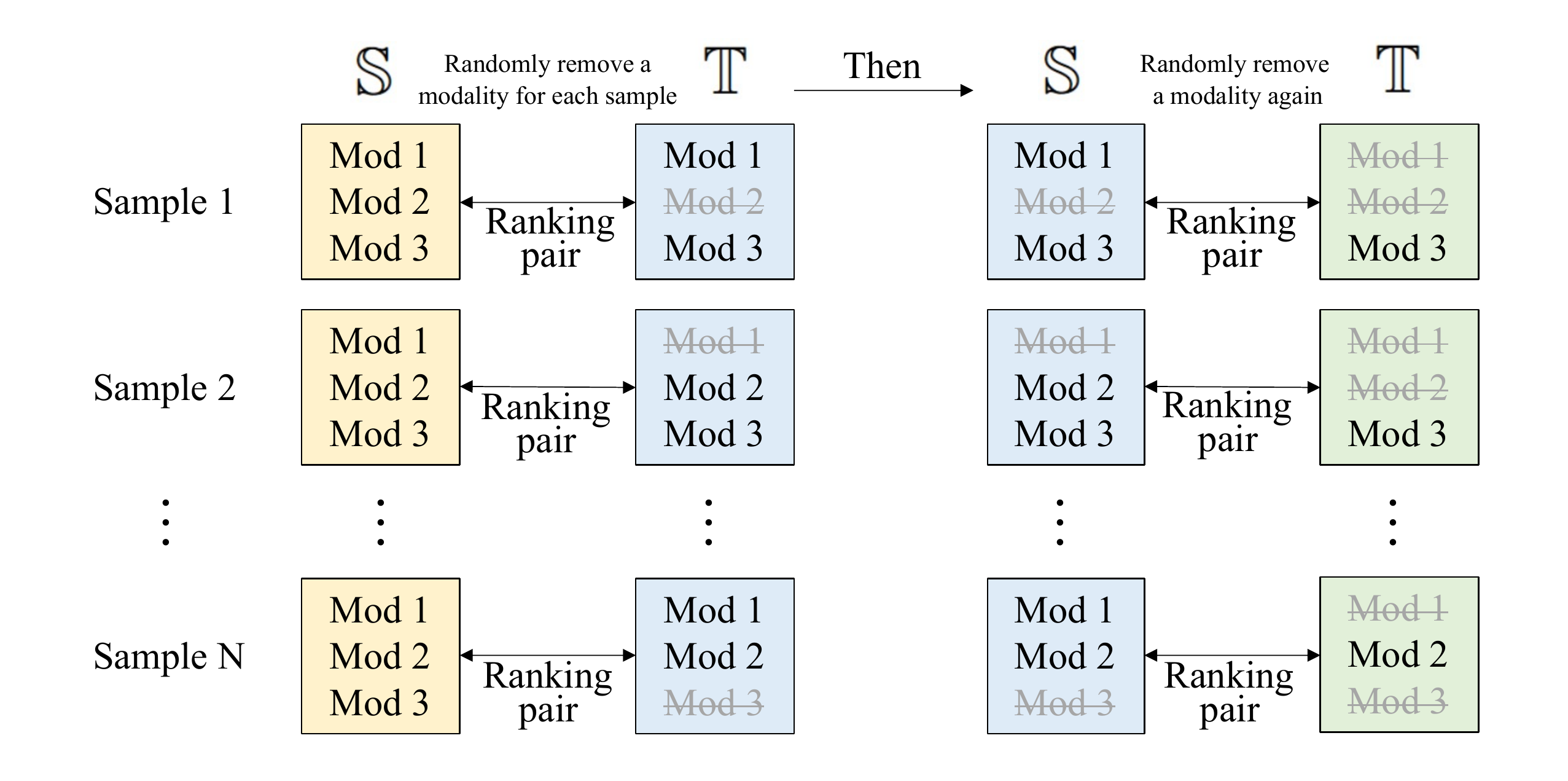}
    \caption{Illustration of generating $\mathbb{S}$ and $\mathbb{T}$.}
    \label{fig:pairs}
\end{figure}

To compute this score in practice, following the prior methods~\cite{moon2020Confidence,toneva2018empirical} we initialize ${\mathbb{S}}$ as the complete modalities, and obtain ${\mathbb{T}}$ by randomly removing a modality from ${\mathbb{S}}$. Then ${\mathbb{T}}$ is regarded as ${\mathbb{S}}$ for another confidence ranking pair and we repeat this process until there is only one modality remained in ${\mathbb{T}}$.

\section{Experiments Details}

\begin{table}[!th]
\vskip 0.15in
\begin{center}
\footnotesize
\caption{Accuracy performance comparison when some of the modalities is blurred (Type I).}\label{tab:app-t1}
\center
\resizebox{\textwidth}{!}{
\begin{tabular}{c|cc|ccccc}
\toprule
\textbf{Dataset}  & \textbf{Noise on}  & \textbf{CML}  &
{\begin{tabular}[c]{@{}c@{}}$\mathbf{\epsilon=0.1}$\\ \end{tabular}}& {\begin{tabular}[c]{@{}c@{}}$\mathbf{\epsilon=0.2}$\\ \end{tabular}}& {\begin{tabular}[c]{@{}c@{}}$\mathbf{\epsilon=0.3}$\\ \end{tabular}}& {\begin{tabular}[c]{@{}c@{}}$\mathbf{\epsilon=0.4}$\\ \end{tabular}}& {\begin{tabular}[c]{@{}c@{}}$\mathbf{\epsilon=0.5}$\\ \end{tabular}}
\\ 
\midrule 
\multirow{17}{*} {\textbf{YaleB}}  
     &  \multirow{2}{*}{\text{\{1\}}}      & \xmark                                                                 &$97.43\pm1.58$   &$96.92\pm1.88$   &$96.41\pm2.20$        &$94.10\pm1.31$       &$92.82\pm1.31$                          \\
                                                                                 &             & \cmark                                                                    &\pmb{$98.46\pm1.09$}  &\pmb{$98.20\pm1.31$}   &\pmb{$96.15\pm1.88$} &\pmb{$94.62\pm1.88$}    &\pmb{$93.59\pm1.30$}                                   \\ \cmidrule{2-8}
      &  \multirow{2}{*}{\text{\{2\}}}       & \xmark                                                              &$95.13\pm0.72$  &$94.10\pm1.31$   &$92.57\pm0.73$   &$92.05\pm1.45$        &$91.54\pm1.66$    \\
                                                                                  
                                                            &                           & \cmark                                                                                                   &\pmb{$96.92\pm1.26$}  &\pmb{$95.90\pm2.02$}
                                       &\pmb{$94.61\pm2.88$} &\pmb{$93.33\pm2.54$}   &\pmb{$93.08\pm3.14$}                \\ \cmidrule{2-8}
     &  \multirow{2}{*}{\text{\{3\}}}      & \xmark                                                                &$94.87\pm0.96$   &$94.87\pm0.96$   &$94.10\pm0.96$        &$92.82\pm1.81$ &$92.05\pm1.31$                                 \\
                                                                                 &             & \cmark                                                                                                      &\pmb{$96.92\pm1.88$}  &\pmb{$97.18\pm1.92$}   &\pmb{$96.15\pm1.88$}   &\pmb{$94.87\pm2.54$}        &\pmb{$94.36\pm2.02$}                       \\ \cmidrule{2-8}
     &  \multirow{2}{*}{\text{\{1, 2\}}}      & \xmark             &$96.67\pm2.61$   &$95.13\pm3.46$   &$91.28\pm2.83$        &$88.72\pm3.10$ &$86.41\pm3.10$                                 \\
                                                                                 &             & \cmark                                                                &\pmb{$97.69\pm0.63$}  &\pmb{$95.39\pm2.26$}   &\pmb{$92.56\pm2.02$}   &\pmb{$89.72\pm2.21$}        &\pmb{$86.66\pm1.81$}                       
                                                                                 \\ \cmidrule{2-8}
     &  \multirow{2}{*}{\text{\{1, 3\}}}      & \xmark          &$97.43\pm0.96$   &$97.69\pm1.66$   &$97.43\pm1.81$        &$97.18\pm2.20$ &$96.15\pm2.26$                                 \\
                                                                                 &             & \cmark                                                    &\pmb{$98.46\pm1.09$}  &\pmb{$98.46\pm1.26$}
                                       &\pmb{$98.46\pm1.66$}
                                       &\pmb{$96.92\pm1.88$}    &\pmb{$96.67\pm2.20$}                                                                          \\
                                                                                 \cmidrule{2-8}
     &  \multirow{2}{*}{\text{\{2, 3\}}}      & \xmark      &$94.62\pm1.08$   &$93.85\pm1.25 $  &$90.26\pm2.54$        &$87.95\pm2.83$ &$86.67\pm2.38$                                 \\
                                                                                 &             & \cmark                       &\pmb{$96.41\pm1.81$}   
                                       &\pmb{$95.64\pm1.92$}   
                                       &\pmb{$93.84\pm3.32$}    
                                       &\pmb{$91.28\pm3.10$} 
                                       &\pmb{$89.49\pm3.16$}                         \\
                                                                                 \cmidrule{2-8}
     &  \multirow{2}{*}{\text{\{1, 2, 3\}}}      & \xmark          &$96.15\pm1.88$  &$96.41\pm3.16$   &$93.85\pm4.40$   &$87.69\pm8.21$        &$84.10\pm10.32$                                \\
                                                                                 &             & \cmark     &\pmb{$97.43\pm1.81$}  &\pmb{$97.43\pm1.92$}   &\pmb{$93.85\pm4.40$}   &\pmb{$87.69\pm7.61$}        &\pmb{$82.56\pm9.26$}
                                                                                                              \\        \midrule 
\multirow{17}{*} {\shortstack{\textbf{Hand-}\\\textbf{written}}}  
     &  \multirow{2}{*}{\text{\{1\}}}      & \xmark          &$97.18\pm1.92$   &$95.38\pm1.25$   &$93.34\pm1.31$        &$92.57\pm1.58$       &$91.28\pm1.31$                          \\
                                                                                 &             & \cmark                                                                                                        &\pmb{$98.46\pm1.26$}  &\pmb{$95.90\pm1.92$}   &\pmb{$93.85\pm1.88$}   &\pmb{$93.08\pm1.66$}        &\pmb{$92.31\pm0.63$}                                        \\ \cmidrule{2-8}
      &  \multirow{2}{*}{\text{\{2\}}}       & \xmark          &$88.46\pm1.66$  &$87.18\pm1.31$   &$86.92\pm1.09$   &$86.92\pm1.09$        &$86.92\pm1.09$    \\
                                                                                  
                                                            &                           & \cmark                                                                                    &\pmb{$90.77\pm3.33$}  &\pmb{$90.26\pm3.57$}   &\pmb{$89.75\pm3.85$}   &\pmb{$89.75\pm3.84$}        &\pmb{$89.75\pm3.84$}                                        \\ \cmidrule{2-8}
     &  \multirow{2}{*}{\text{\{3\}}}      & \xmark          &$85.90\pm1.92$   &$85.13\pm1.81$   &$84.87\pm1.45$        &$84.62\pm1.66$ &$84.62\pm1.66$                                 \\
                                                                                 &             & \cmark                                                                                                        &\pmb{$88.97\pm2.54$}  &\pmb{$88.21\pm2.61$}   &\pmb{$87.69\pm2.74$}   &\pmb{$87.69\pm3.32$}        &\pmb{$87.44\pm3.10$}                       \\ \cmidrule{2-8}
     &  \multirow{2}{*}{\text{\{1, 2\}}}      & \xmark         &$88.97\pm3.68$   &$83.08\pm3.50$   &$78.97\pm1.92$        &$77.69\pm2.74$ &$75.90\pm3.57$                                 \\
                                                                                 &             & \cmark                                                     &\pmb{$88.97\pm4.04$}  &\pmb{$83.59\pm2.97$}   &\pmb{$80.51\pm3.46$}   &\pmb{$77.18\pm4.28$}        &\pmb{$74.10\pm3.84$}                       
                                                                                 \\ \cmidrule{2-8}
     &  \multirow{2}{*}{\text{\{1, 3\}}}      & \xmark          &$91.54\pm1.09$   &$91.28\pm3.16$   &$88.97\pm5.41$        &$87.43\pm5.83$ &$85.64\pm6.42$                                 \\
                                                                                 &             & \cmark                                                   &\pmb{$93.59\pm2.38$}  &\pmb{$91.79\pm3.68$}
                                        &\pmb{$88.97\pm4.04$}
                                     &\pmb{$86.93\pm4.99$}   &\pmb{$85.39\pm4.91$}                                                                          \\
                                                                                 \cmidrule{2-8}
     &  \multirow{2}{*}{\text{\{2, 3\}}}      & \xmark          &$63.59\pm8.00$   &\pmb{$59.74\pm7.00$}   &\pmb{$57.69\pm5.99$}        &\pmb{$56.67\pm5.94$} &\pmb{$55.90\pm5.49$}                                 \\
                                                                                 &             & \cmark                                                              &\pmb{$64.36\pm7.49$}   &$58.46\pm6.37$   &$56.67\pm6.10$        &$55.64\pm6.04$ &$54.87\pm6.29$                                                                  \\
                                                                                 \cmidrule{2-8}
     &  \multirow{2}{*}{\text{\{1, 2, 3\}}}      & \xmark          &$54.87\pm10.68$  &\pmb{$37.95\pm6.92$}   &\pmb{$29.48\pm4.76$}   &\pmb{$24.36\pm4.04$}        &\pmb{$22.31\pm4.12$}                                \\
                                                                                 &             & \cmark                                                     &\pmb{$57.18\pm11.41$}  &$35.64\pm4.80$   &$26.67\pm2.54$   &$22.82\pm2.54$        &$20.77\pm1.09$
                                                                                                              \\ 
   \midrule 
\multirow{7}{*} {\shortstack{\textbf{TUAND-}\\\textbf{ROMD}}}  
     &  \multirow{2}{*}{\text{\{1\}}}      & \xmark          &$84.77\pm0.55$                                                    &$80.47\pm0.99$  &$76.53\pm1.11$   &$72.65\pm0.76$   &$70.17\pm0.66$                                          \\
                                                                                 &             & \cmark               &\pmb{$86.50\pm0.59$}               &\pmb{$82.46\pm0.77$}   &\pmb{$78.30\pm1.18$}   &\pmb{$74.92\pm1.39$}        &\pmb{$72.45\pm1.33$}                                        \\ \cmidrule{2-8}
      &  \multirow{2}{*}{\text{\{2\}}}       & \xmark                                                              &$86.56\pm0.27$  &$85.71\pm0.48$   &$84.14\pm0.58$   &$82.35\pm0.86$        &$80.85\pm1.05$    \\
                                                                                  
                                                            &                           & \cmark                                                                                   &\pmb{$88.87\pm0.22$}               &\pmb{$88.74\pm0.28$}   &\pmb{$88.58\pm0.63$}   &\pmb{$88.15\pm0.65$}        &\pmb{$87.93\pm0.67$}                                          \\ \cmidrule{2-8}
     &  \multirow{2}{*}{\text{\{1, 2\}}}      & \xmark          &$84.88\pm1.19$                                                    &$80.72\pm1.02$  &$76.60\pm0.75$  &$73.15\pm1.10$  &$70.35\pm1.25$                                            \\
                                                                                 &             & \cmark                                                    &\pmb{$87.41\pm3.40$}                                                    &\pmb{$82.78\pm1.14$}  &\pmb{$79.28\pm1.00$}   &\pmb{$76.30\pm1.11$}   &\pmb{$73.82\pm1.35$}                                                                                                             \\
                                                                             \bottomrule
\end{tabular}}

\end{center}
\end{table}

\begin{table}[!th]
\vskip 0.15in
\begin{center}
\footnotesize
\caption{Accuracy performance comparison when some of the modalities is blurred (Type II).}\label{tab:app-t2}
\center
\resizebox{\textwidth}{!}{
\begin{tabular}{c|cc|ccccc}
\toprule
\textbf{Dataset}  & \textbf{Noise Noise on}  & \textbf{CML}  &
{\begin{tabular}[c]{@{}c@{}}$\mathbf{\epsilon=0.5}$\\ \end{tabular}}& {\begin{tabular}[c]{@{}c@{}}$\mathbf{\epsilon=1.0}$\\ \end{tabular}}& {\begin{tabular}[c]{@{}c@{}}$\mathbf{\epsilon=1.5}$\\ \end{tabular}}& {\begin{tabular}[c]{@{}c@{}}$\mathbf{\epsilon=2.0}$\\ \end{tabular}}& {\begin{tabular}[c]{@{}c@{}}$\mathbf{\epsilon=2.5}$\\ \end{tabular}}
\\ \midrule
\multirow{17}{*} {\textbf{YaleB}}  
     &  \multirow{2}{*}{\text{\{1\}}}      & \xmark   &$ 95.90\pm2.54 $   &$ 94.87\pm3.22 $        &$ 93.85\pm2.88 $       &$ 93.59\pm3.16 $       &$ 93.59\pm3.16 $                   \\
   &             & \cmark           &\pmb{$ 97.43\pm1.31 $}  &\pmb{$ 96.15\pm2.51 $}   &\pmb{$ 95.13\pm2.97 $}   &\pmb{$ 94.36\pm2.97 $}        &\pmb{$ 93.85\pm3.46 $}               \\ \cmidrule{2-8}
      &  \multirow{2}{*}{\text{\{2\}}}       & \xmark          &$ 96.15\pm2.26 $  &$ 93.33\pm3.22 $   &$ 91.03\pm2.62 $   &$ 90.26\pm2.02 $        &$ 89.23\pm2.18 $    \\  &                           & \cmark                                                                                    &\pmb{$ 97.69\pm1.26 $}  &\pmb{$ 96.67\pm1.58 $}   &\pmb{$ 94.10\pm2.20 $}   &\pmb{$ 92.82\pm2.83 $}        &\pmb{$ 92.05\pm2.02 $}                                        \\ \cmidrule{2-8}
     &  \multirow{2}{*}{\text{\{3\}}}      & \xmark &$ 98.72\pm0.36 $   &$ 96.92\pm1.26 $   &$ 96.15\pm0.63 $        &$ 96.15\pm0.63 $ &$ 95.90\pm0.96 $                                 \\      &             & \cmark                                                    &\pmb{$ 98.72\pm0.73 $}  &\pmb{$ 97.69\pm1.09 $}   &\pmb{$ 97.43\pm0.96 $}   &\pmb{$ 97.18\pm1.31 $}        &\pmb{$ 96.67\pm1.58 $}                       \\ \cmidrule{2-8} &  \multirow{2}{*}{\text{\{1, 2\}}}      & \xmark          &$ 95.64\pm2.83 $   &$ 91.02\pm3.46 $        &$ 88.46\pm4.53 $ &$ 87.18\pm3.46 $      &$ 85.90\pm4.09 $                           \\                        &             & \cmark                                                     &\pmb{$ 96.66\pm1.31 $}   &\pmb{$ 93.59\pm2.38 $}        &\pmb{$ 90.51\pm2.97 $} &\pmb{$ 86.67\pm3.46 $}      &\pmb{$ 84.62\pm3.26 $}   \\ \cmidrule{2-8}
     &  \multirow{2}{*}{\text{\{1, 3\}}}      & \xmark          &$ 98.46\pm0.63 $   &$ 98.46\pm1.66 $   &$ 97.69\pm1.66 $        &$ 97.43\pm1.45 $ &$ 97.18\pm1.31 $                                 \\                  &             & \cmark                                                     &\pmb{$ 98.20\pm0.73 $}  &\pmb{$ 97.95\pm1.92 $}   &\pmb{$ 97.69\pm1.66 $}     &\pmb{$ 98.20\pm1.58 $}   &\pmb{$ 97.69\pm1.66 $}     \\                    \cmidrule{2-8}   &  \multirow{2}{*}{\text{\{2, 3\}}}      & \xmark          &$ 97.43\pm0.36 $   &$ 95.89\pm0.36 $   &$ 95.38\pm0.62 $        &$ 94.62\pm0.62 $ &$ 92.82\pm0.73 $                                 \\         &             & \cmark                                                    &\pmb{$ 98.72\pm0.36 $}   &\pmb{$ 97.69\pm1.09 $}   &\pmb{$ 96.66\pm0.73 $}        &\pmb{$ 95.38\pm0.62 $} &\pmb{$ 94.61\pm1.66 $}    \\    \cmidrule{2-8}   &  \multirow{2}{*}{\text{\{1, 2, 3\}}}      & \xmark          &$ 97.69\pm0.63 $  &$ 95.64\pm0.36 $   &$ 93.08\pm1.09 $   &$ 89.23\pm1.66 $        &$ 82.31\pm1.26 $                                \\   &             & \cmark                              &\pmb{$ 98.46\pm0.63 $}  &\pmb{$ 97.18\pm1.31 $}   &\pmb{$ 95.64\pm0.96 $}   &\pmb{$ 92.56\pm2.54 $}        &\pmb{$ 88.46\pm2.27 $}  \\   \midrule

\multirow{7}{*} {\textbf{CUB}}  
     &  \multirow{2}{*}{\text{\{1\}}}      & \xmark &$ 91.11\pm1.04 $  &$ 86.94\pm2.83 $   &$ 83.61\pm3.93 $   &$ 80.83\pm4.14 $        &$ 79.17\pm3.79 $                                  \\
                                                                                 &             & \cmark   &\pmb{$ 93.33\pm1.80 $}   &\pmb{$ 90.83\pm2.45 $}   &\pmb{$ 87.50\pm3.60 $}        &\pmb{$ 85.56\pm4.38 $}  &\pmb{$ 81.11\pm4.53 $}                                      \\ \cmidrule{2-8}
      &  \multirow{2}{*}{\text{\{2\}}}       & \xmark          &$ 91.11\pm0.40 $  &$ 91.95\pm0.39 $   &$ 91.11\pm0.40 $   &$ 89.72\pm0.39 $        &$ 88.61\pm0.79 $    \\
                                                                                  
                                                            &                           & \cmark                                                                                   &\pmb{$ 93.61\pm1.04 $}   &\pmb{$ 92.78\pm1.04 $}   &\pmb{$ 92.50\pm1.80 $}        &\pmb{$ 91.67\pm2.96 $}  &\pmb{$ 91.39\pm3.22 $}                                        \\ \cmidrule{2-8}
     &  \multirow{2}{*}{\text{\{1, 2\}}}      & \xmark          &$ 92.78\pm1.97 $  &$ 88.61\pm1.42 $  &$ 85.83\pm1.80 $   &$ 79.72\pm2.83 $      &$ 74.17\pm4.46 $                                   \\
                                                                                 &             & \cmark                                                     &\pmb{$ 94.72\pm2.19 $}  &\pmb{$ 92.22\pm3.75 $}   &\pmb{$ 90.00\pm4.46 $}   &\pmb{$ 86.11\pm4.10  $}       &\pmb{$ 79.17\pm4.91 $}                                                   \\
                                                                                 \midrule
\multirow{7}{*} {\textbf{Animal}}  
     &  \multirow{2}{*}{\text{\{1\}}}      & \xmark          &$ 86.61\pm0.20 $  &$ 85.81\pm0.36 $   &$ 84.82\pm1.02 $   &$ 83.77\pm1.29 $        &$ 82.16\pm2.32 $                                  \\
                                                                                 &             & \cmark                &\pmb{$ 87.20\pm0.18 $}   &\pmb{$ 87.01\pm0.18 $}   &\pmb{$ 86.60\pm0.20 $}        &\pmb{$ 86.03\pm0.04 $}  &\pmb{$ 85.42\pm0.29 $}                                      \\ \cmidrule{2-8}
      &  \multirow{2}{*}{\text{\{2\}}}       & \xmark          &$ 86.33\pm0.54 $  &$ 85.62\pm0.61 $   &$ 84.84\pm0.95 $   &$ 83.04\pm1.24 $        &$ 81.34\pm1.73 $    \\
                                                                                  
                                                            &                           & \cmark                                                                                   &\pmb{$ 87.04\pm0.08 $}   &\pmb{$ 86.64\pm0.26 $}   &\pmb{$ 85.95\pm0.42 $}        &\pmb{$ 84.78\pm0.17 $}  &\pmb{$ 82.71\pm0.24 $}                                        \\ \cmidrule{2-8}
     &  \multirow{2}{*}{\text{\{1, 2\}}}      & \xmark          &$ 86.01\pm0.17 $  &$ 84.80\pm0.81 $  &$ 83.17\pm1.65 $  &$ 80.92\pm2.77 $   &$ 77.42\pm4.14 $                                         \\
                                                                                 &             & \cmark                                                     &\pmb{$ 87.04\pm0.42 $}  &\pmb{$ 86.50\pm0.15 $}   &\pmb{$ 85.38\pm0.34 $}   &\pmb{$ 83.84\pm0.65 $}       &\pmb{$ 81.67\pm0.75 $}                                                                                                      \\ 
                                                                                 \midrule
\multirow{7}{*} {\shortstack{\textbf{TUAND-}\\\textbf{ROMD}}}  
     &  \multirow{2}{*}{\text{\{1\}}}      & \xmark          &$ 81.14\pm0.70 $                                                    &$ 78.21\pm0.92 $  &$ 75.39\pm1.09 $   &$ 73.21\pm1.46 $   &$ 71.71\pm1.26 $                                          \\
                                                                                 &             & \cmark                            &\pmb{$ 81.99\pm1.99 $}   &\pmb{$ 78.79\pm2.42 $}   &\pmb{$ 76.37\pm2.57 $}        &\pmb{$ 74.36\pm2.63 $}      &\pmb{$ 73.19\pm2.60 $}
                                                                                 \\ \cmidrule{2-8}
      &  \multirow{2}{*}{\text{\{2\}}}       & \xmark                                                              &$ 84.19\pm0.82 $  &$ 84.43\pm0.48 $  &$ 84.46\pm0.35 $ &$ 84.32\pm0.45 $   &$ 84.21\pm0.44 $            \\
                                                                                  
                                                            &                           & \cmark                                                                                   &\pmb{$ 84.88\pm1.62 $}               &\pmb{$ 84.73\pm1.89 $}   &\pmb{$ 84.84\pm1.76 $}   &\pmb{$ 84.39\pm0.89 $}        &\pmb{$ 84.97\pm1.52 $}                                          \\ \cmidrule{2-8}
     &  \multirow{2}{*}{\text{\{1, 2\}}}      & \xmark          &$ 83.56\pm1.23 $                                                    &$ 80.85\pm1.30 $  &$ 77.85\pm1.53 $  &$ 75.90\pm2.07 $  &$ 74.08\pm2.22 $                                            \\
                                                                                 &             & \cmark                                                    &\pmb{$ 83.99\pm1.87 $}                 &\pmb{$ 81.48\pm2.30 $}  &\pmb{$ 78.50\pm2.30 $}   &\pmb{$ 76.73\pm2.19 $}   &\pmb{$ 75.23\pm2.20 $}                                                                                                             \\ 
                                                    \bottomrule
\end{tabular}}
\end{center}
\end{table}
\subsection{Dataset Details}\label{sec:app-datasets}
We evaluate the proposed method on diverse datasets, including data with multiple modalities and multiple types of features. $\circ$~\textbf{YaleB}: Similar to previous work~\cite{geo2002From}, we also use a subset of this face image dataset, which contains $650$ facial images, $10$ classes and $3$ different types of features. $\circ$~\textbf{Handwritten}~\cite{perkins2003Online}: This is a database of handwritten digits which contains $2,000$ images, $10$ classes, $6$ types of features. $\circ$~\textbf{CUB}~\cite{wah2011caltech}: Following CPM-Nets~\cite{zhang2019cpm}, we use a subset of this dataset, which contains first $10$ classes of original dataset and $2$ modalities (deep visual feature and text feature) are obtained by GoogleNet and doc2vec~\cite{le2014distributed}. $\circ$~\textbf{Animal}: This dataset contains $10,158$ images, $50$ classes, and $2$ types of features (deep visual feature from DECAF~\cite{krizhevsky2012imagenet} and VGG19~\cite{simonyan2014very}). $\circ$ \textbf{TUANDROMD}~\cite{borah2020malware}: The dataset contains $4,465$ instances, $2$ classes and $2$ types of modalities.

\subsection{Experiment Setting}\label{sec:app-setting}
\textbf{Type-I}: For CPM-Nets and the first five datasets(i.e.,YaleB, Handwritten, CUB and Animal), we follow the author's implementation~\cite{zhang2019cpm}: the dimensionality of latent representation is $150$. Parameter lambda for cub/animal/hand-written/yaleB/tuandromd is set as 5/45/45/10/5. The dimensionalities of input, hidden layers are $128$ and $300$. We use Adam optimizer to train all CPM-Nets models with the learning rate of $10^{-2}$ and no additional regularization term. For Tuandromd dataset, we tune the dimensionality of latent representation to $512$. The dimensionalities of input and hidden layers are both $512$. We use Adam optimizer to train CPM-Net with L2-regularization term.
\textbf{Type-II}: For MIWAE, we train the encoder, decoder and classifier respectively. The number of hidden units of them is all $128$. Parameter lambda for cub/animal/hand-written/yaleB/tuandromd are set as 15/25/10/35/75 for best performance. The dimensionalities of the latent space are $64$. We use Adam optimizer to train the encoder and decoder with a learning rate of $10^{-2}$. Then we train the encoder, decoder and classifier altogether for another with a learning rate of $10^{-3}$. As same as prior work~\cite{2020address}, we evaluated the performance according to Accuracy ($\%$), NLL ($10^{-1}$), AURC ($10^{-3}$), and E-AURC ($10^{-3}$).

\subsection{Robustness Evaluation}\label{sec:app-noise}

We evaluate models in terms of accuracy under Gaussian noise (i.e., zero mean and varying variance $\epsilon$), and ``Noise On'' indicates which modality is noised (e.g., $\{1\}$ indicates the first modality is noised). In addition to the performance on the challenging datasets (CUB and Animal) in the main text (Table~\ref{tab:noise}), we show more other results (Table~\ref{tab:app-t1}~\ref{tab:app-t2}). It is clear that the models equipped with CML are more robust to noise, especially when the noise is much heavier.

\subsection{Additional Results for Robustness Estimation}\label{sec:app-acc}
\begin{table}[!ht]
\vskip 0.15in
\begin{center}
\caption{Accuracy performance comparison for whether the model is equipped with the cma regularization term on additional dataset (i.e., whether $\lambda$ is set to 0).}
\label{tab:additionalacc}
\center
\scalebox{1.00}{
\setlength{\tabcolsep}{1.75mm}\begin{tabular}{c|c|ccccc}
\toprule
\multicolumn{1}{c}{\text{Method}}  & \multicolumn{1}{c}{\text{Dataset}}  & \text{CML}  & \text{\begin{tabular}[c]{@{}c@{}}Accuracy\\ ($\uparrow$)\end{tabular}} & \text{\begin{tabular}[c]{@{}c@{}}NLL\\ ($\downarrow$)\end{tabular}} & \text{\begin{tabular}[c]{@{}c@{}}AURC\\ ($\downarrow$)\end{tabular}}& \text{\begin{tabular}[c]{@{}c@{}}E-AURC\\ ($\downarrow$)\end{tabular}} \\ \midrule 
\multirow{6}{*} {\text{Type I}}     &  \multirow{3}{*}{\text{YaleB}}       & \xmark          &$95.84\pm0.78$                                                    &$21.98\pm0.05$  &$3.00\pm1.38$   &$2.08\pm1.37$             \\
                                                                                  
                                                            &                           & \cmark                                                                                   &${97.69\pm1.09}$                                                    &$21.98\pm0.05$  &${1.46\pm1.51}$   &${1.12\pm1.32}$  \\& &  Improve & \textcolor{mycolor2}{{$~~~~~~~~\bigtriangleup 1.85 $}} & \textcolor{gray}{$~~~~~~~~~~~0.00$}& \textcolor{mycolor2}{{$~~~~~~\bigtriangleup 1.54 $}}& \textcolor{mycolor2}{{$~~~~~~~\bigtriangleup 0.96 $}}                                               \\ \cmidrule{2-7}
      &  \multirow{3}*{\shortstack{\text{Hand-}\\\text{written}}}      & \xmark          &$89.00\pm3.64$                                                    &$20.30\pm0.25$  &$35.83\pm20.43$   &$28.80\pm15.49$          \\
                                                                                  
                                                            &                           & \cmark                                                                                   &${93.60\pm0.60}$                                                    &${20.06\pm0.11}$  &${11.00\pm6.17}$   &${8.90\pm5.80}$                  \\& & Improve & \textcolor{mycolor2}{{$~~~~~~~~\bigtriangleup 4.60 $}} & \textcolor{mycolor2}{{$~~~~~~~~\bigtriangleup 0.14 $}}& \textcolor{mycolor2}{{$~~~~~~\bigtriangleup 14.83 $}}& \textcolor{mycolor2}{{$~~~~~~~\bigtriangleup 19.90 $}}                  
    
                                                           \\ \midrule
\multirow{6}{*} {\text{Type II}}    &  \multirow{3}{*}{\text{YaleB}}       & \xmark          &$95.69\pm2.10$                                                    &$1.80\pm0.71$  &$5.50\pm2.86$   &${4.32\pm2.32}$            \\
                                                                                  
                                                            &                           & \cmark                                                                                   &${97.84\pm0.58}$                                                    &${1.11\pm0.49}$  &${5.02\pm6.39}$   &$4.76\pm6.26$       \\& & Improve & \textcolor{mycolor2}{{$~~~~~~~~\bigtriangleup 2.15 $}} & \textcolor{mycolor2}{{$~~~~~~\bigtriangleup 0.69 $}}& \textcolor{mycolor2}{{$~~~~~~\bigtriangleup 0.48 $}}& \textcolor{mycolor1}{{$~~~~~~~\bigtriangledown 0.44 $}}                                           \\ \cmidrule{2-7}
      &  \multirow{3}*{\shortstack{\text{Hand-}\\\text{written}}}      & \xmark          &$98.40\pm0.64$                                                  &$0.49\pm0.12$  &$0.32\pm0.16$   &$0.16\pm0.12$                                          \\
                                                                                 &             & \cmark                                                    &${99.05\pm0.19}$                                                    &$0.50\pm0.10$  &${0.18\pm0.07}$   &${0.14\pm0.08}$                          \\& & Improve & \textcolor{mycolor2}{{$~~~~~~~~\bigtriangleup 0.65 $}} & \textcolor{gray}{$~~~~~~~~~~~0.00$}& \textcolor{mycolor2}{{$~~~~~~\bigtriangleup 0.14 $}}& \textcolor{mycolor2}{{$~~~~~~~\bigtriangleup 0.02 $}}             
     
 \\   \bottomrule
\end{tabular}}
\end{center}
\end{table}

Limited by space, we show the performance of model equipped with CML on YaleB and Handwritten. 
From Table \ref{tab:additionalacc}, the classification models
equipped with CML consistently outperforms their counterpart validating the rationality of CML principle.

\subsection{Confidence Estimation for Complete Inputs}\label{sec:app-conf}

\begin{figure}[!ht] 
  \centering
  \subfloat[CPM-Nets (Complete)]{
  \includegraphics[width=0.3\textwidth]{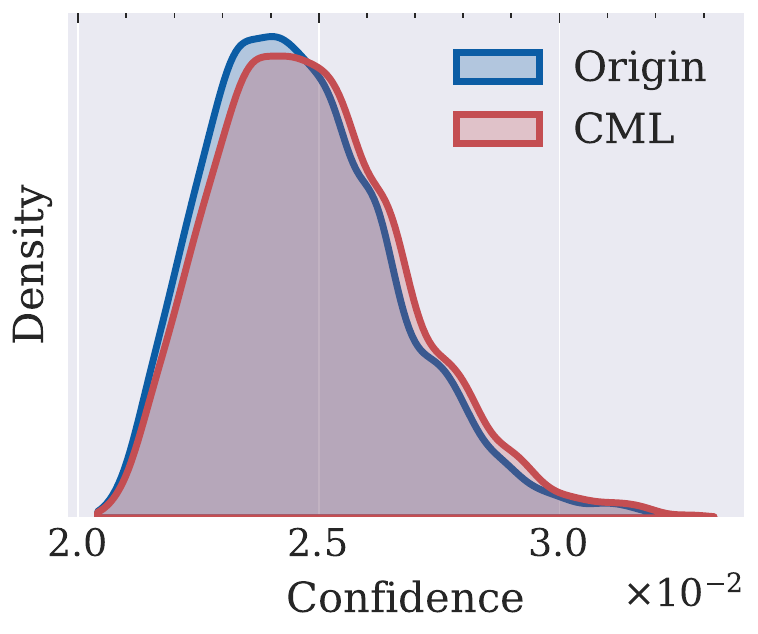}
  }
    \subfloat[MIWAE (Complete)]{
  \includegraphics[width=0.3\textwidth]{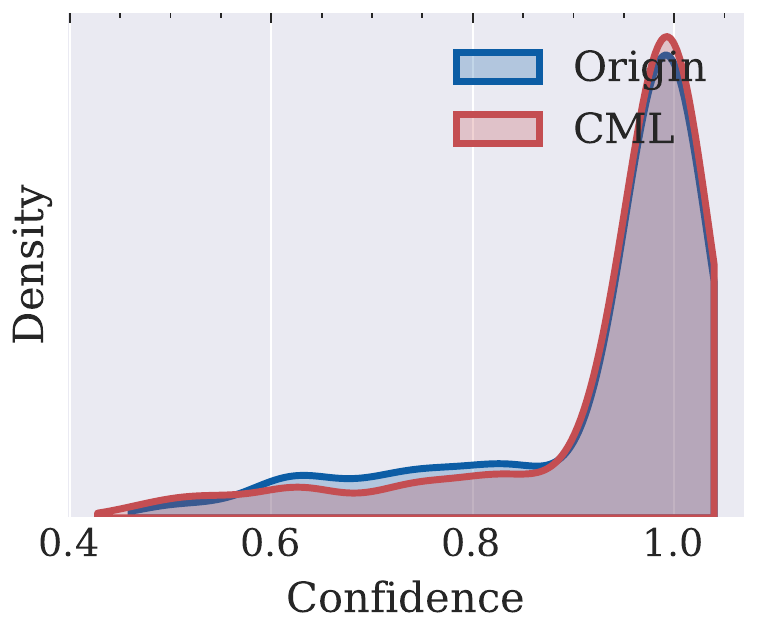}
  }
  \subfloat[MMTM (Complete)]{
  \includegraphics[width=0.3\textwidth]{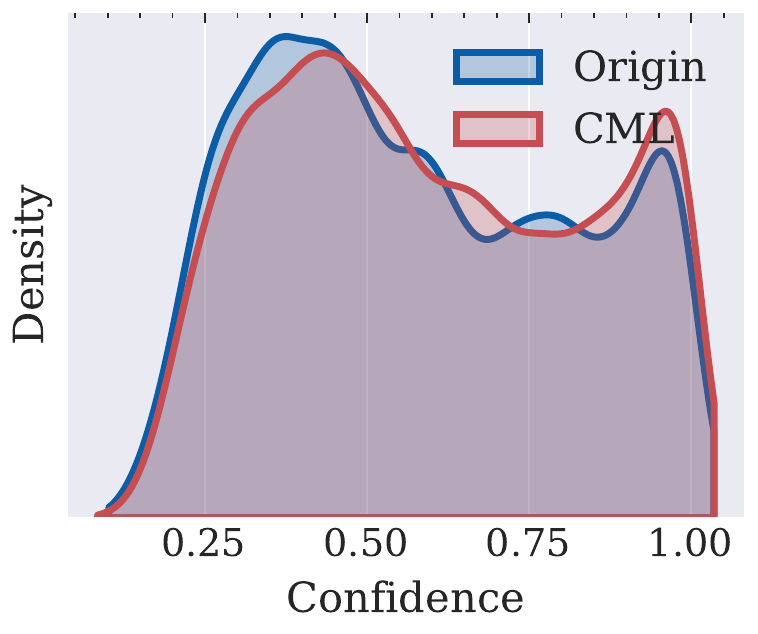}
  } \\
  \subfloat[CPM-Nets (Removed)]{
  \includegraphics[width=0.3\textwidth]{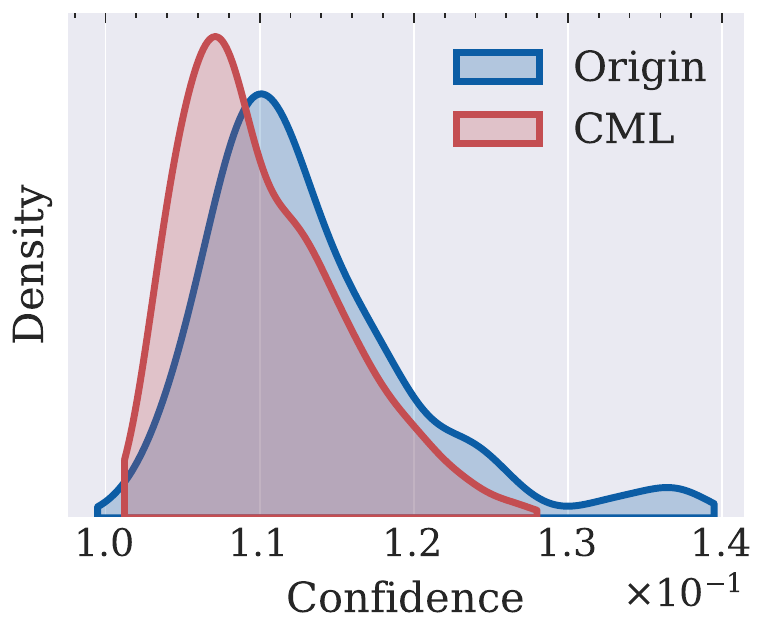}
  }
    \subfloat[MIWAE (Removed)]{
  \includegraphics[width=0.3\textwidth]{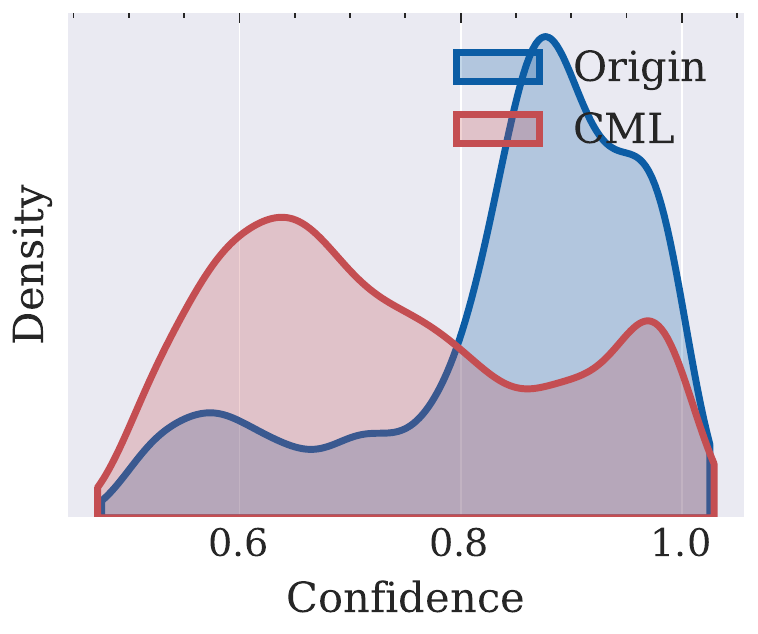}
  }
  \subfloat[MMTM (Removed)]{
  \includegraphics[width=0.3\textwidth]{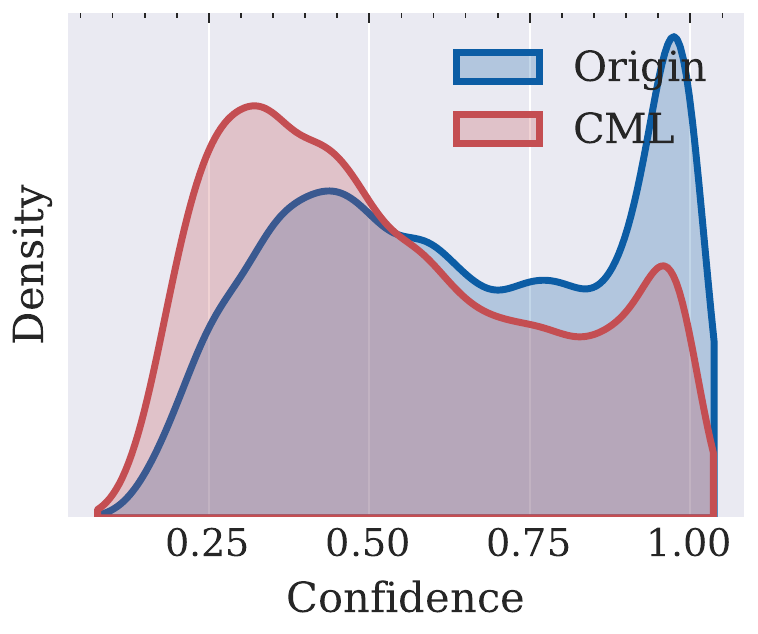}
  }
 \caption{Confidence estimation on complete inputs. We estimate the confidence on complete inputs (top) and the confidence when one modality is removed (bottom). We can find CML regularization keeps the confidence estimation on complete input but alleviate the over-confidence when one modality is removed, which indicates the proposed method calibrates the multimodal model by rethinking the relationship between the modalities.}
  \label{fig:conf}
\end{figure}

We show the confidence estimation for complete inputs, as shown in Fig.~\ref{fig:conf}, we can find that the confidence estimation of original model and CML model are very similar. To prevent the model from being over-confident when model predicts a wrong prediction, the regularization will not be added when prediction of complete input is wrong. From the bottom figures, we can find CML regularization alleviates the problem that model increases the confidence when one modality is removed.  

\textbf{Proof} of Lemma~\ref{lem:3-2}: if we have $\mathrm{VRR}_{CML} < \mathrm{VRR}_\text{ORIG}$, then we have $\mathbb{E}\left(\mathrm{Conf}_{CML}({x}^{(\mathbb{T})})\right) - \mathbb{E}\left(\mathrm{Conf}_{CML}({x}^{(\mathbb{S})})\right) \leq \mathbb{E}\left(\mathrm{Conf}_\text{ORIG}({x}^{(\mathbb{T})})\right)-\mathbb{E}\left(\mathrm{Conf}_\text{ORIG}({x}^{(\mathbb{S})})\right)$, then we have:
\begin{equation}
\begin{aligned}
    &\mathbb{E}\left(\mathrm{Conf}_{CML}({x}^{(\mathbb{T})})\right)\leq\mathbb{E}\left(\mathrm{Conf}_\text{ORIG}({x}^{(\mathbb{T})})\right), \\ &\text{subject to: } \mathbb{E}\left(\mathrm{Conf}_{CML}({x}^{(\mathbb{T})})\right)=\mathbb{E}\left(\mathrm{Conf}_{ORIG}({x}^{(\mathbb{T})})\right)
\end{aligned}
\end{equation}

During the train stage, we evaluate the confidence difference between the $\mathbb{E}\left(\mathrm{Conf}_{CML}({x}^{(\mathbb{T})})\right)$ and $\mathbb{E}\left(\mathrm{Conf}_{ORIG}({x}^{(\mathbb{T})})\right)$, i.e., $\mathbb{E}\left(\left|\mathrm{Conf}_{CML}({x}^{(\mathbb{T})})-\mathrm{Conf}_{ORIG}({x}^{(\mathbb{T})})\right|\right)$. We find the confidence difference between the $\mathbb{E}\left(\mathrm{Conf}_{CML}({x}^{(\mathbb{T})})\right)$ and $\mathbb{E}\left(\mathrm{Conf}_{ORIG}({x}^{(\mathbb{T})})\right)$ is very small (less than $0.1\%$), which implies that the confidence estimation on complete inputs are very close.

\subsection{Confidence Estimation when Just Penalizing the Confidence Difference}\label{sec:app-diff}

\begin{figure}[!ht] 
  \centering
  \subfloat[Modality 1]{
  \includegraphics[width=0.3\textwidth]{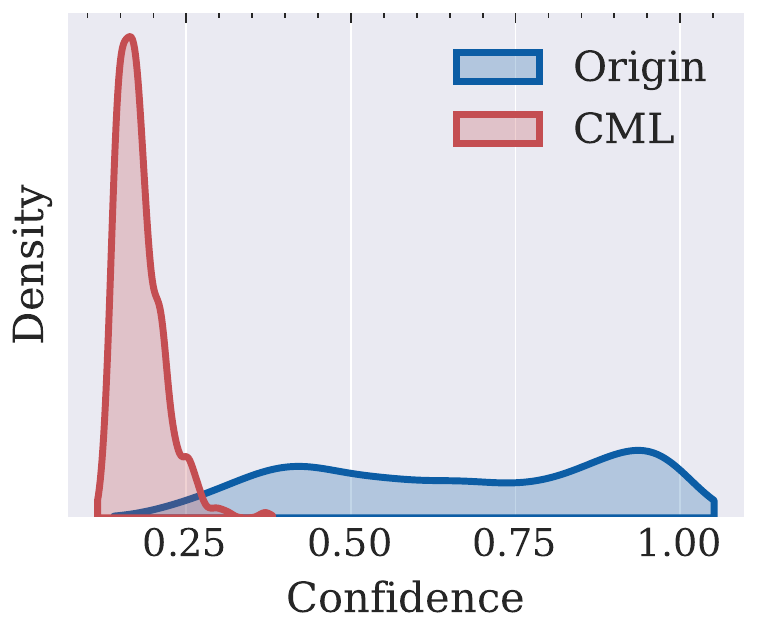}
  }
    \subfloat[Modality 2]{
  \includegraphics[width=0.3\textwidth]{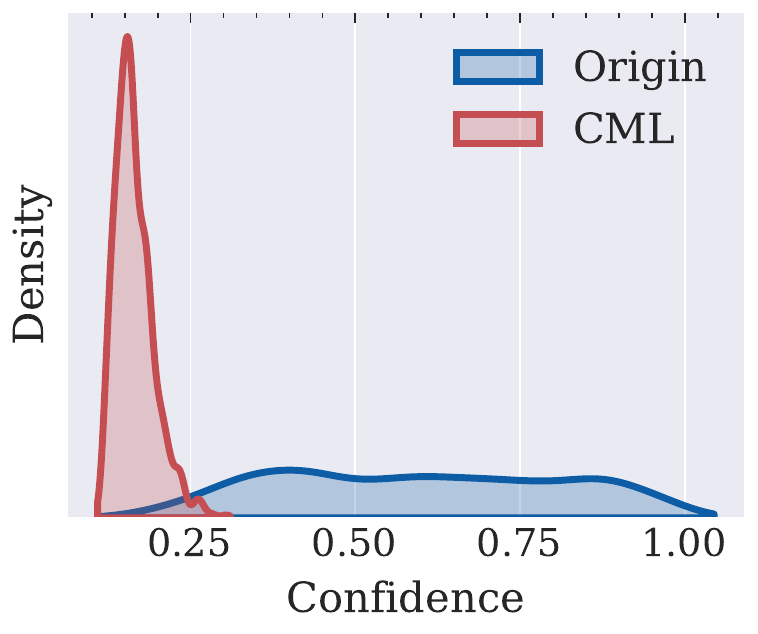}
  }
  \subfloat[Complete modalities]{
  \includegraphics[width=0.3\textwidth]{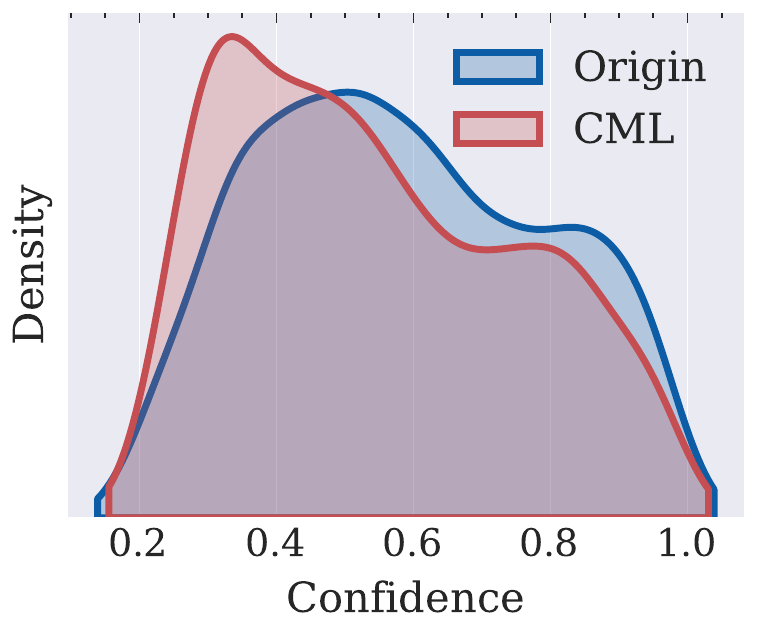}
  }

 \caption{Confidence estimation when penalizing the confidence difference (Eq.~\ref{eq:Ldiff}).}
  \label{fig:app-diff}
\end{figure}

Forcing the confidence for ${x}^{({\mathbb{T}})}$ to be smaller than the confidence for ${x}^{({\mathbb{S}})}$ strictly (Eq.~\ref{eq:Ldiff}) will lead to a very small confidence for ${x}^{({\mathbb{T}})}$ and will make the model estimate an extremely small confidence for each modality, which contradicts the fact that the model sometimes can still make correct predictions confidently when one modality is removed. A flexible ranking regularization makes it more suitable for real data.

\section{Analysis of the Training Time and Space Complexity}
Ideally, CML should be computed over all possible pairs at each model update. However, it is computationally expensive, so we employ an approximation scheme following \cite{toneva2018empirical} for reducing the costs. For example, given samples with 4 modalities (a, b, c, d), we need to sample 3 pairs (a/ab, ab/abc, abc/abcd) to approximate CML loss, and indexes are shuffled for different epochs. So if the complexity of the traditional model is o(n), the complexity of our method will be o((k-1)n), where k indicates the number of modalities. It should be pointed out that compared models in our experiments are also equipped with sampling (to avoid the influence of sampling), and the complexity of compared methods is also o((k-1)n). We report the training time (seconds) for the same training epochs (Platform: RTX 3090$\times$8, CUDA Version: 11.2). It is observed that the original model and model equipped with CML have the same level of computational complexity.

\begin{table}[ht]

\begin{center}
\caption{Training time (Platform: RTX 3090 $\times8$).}
\label{tab:time}
\center
 \scalebox{1.00}{
 \setlength{\tabcolsep}{1.3mm}
 \begin{tabular}{c|ccccccc}
\toprule
\multicolumn{1}{c}{\text{Method}}   & \text{CML}   & \text{TUANDROMD} & \text{YaleB}& \text{Handwritten} & \text{CUB} &\text{Animal} \\ \midrule
\multirow{2}{*} {$\text{Type I}$}   & \xmark    &  $245.3$                                                    &$1574.6$  &$141.5$   &$351.6$   &$1582.7$                                                     \\
  
        & \cmark                                                                                   &$297.6$                                                    &$1210.2$  &$191.2$   &${348.5}$   &${1641.3}$     \\ \midrule 
\multirow{2}{*}{{$\text{Type II}$}}
     & \xmark       &$1447.7$ &$703.3$    &$233.2$     &$565.2$    &$717.8$                    \\
     & \cmark                                                    &${1489.1}$   &${662.9}$  &${210.8}$  &${781.7}$  &${  720.3}$     \\ \bottomrule
\end{tabular}}
\end{center}
\end{table}

\section{Algorithms}
In addition to the general algorithm shown in the main text, we show the specific algorithms corresponding to different types of algorithms and add more comments for better understanding. 
\subsection{CML for Imputation-independent Model}
\begin{algorithm}[ht]
 	\caption{CML for the imputation-independent model}
 	 	\label{alg:cpm}
    \begin{algorithmic}
 	\STATE	\textbf{Given} dataset $\mathcal{D}=\left\{\{{x}_i^m\}_{m=1}^M,y_i\right\}_{i=1}^N$, classifier $f$, and classification loss function $\mathcal{L}^\text{CL}$, Coefficient $\lambda$ of CML, epochs for training the classifier ${epoch}$
 		
 		\FOR{$e=1,\ldots,{epoch}$}{
 		\STATE $\mathbb{S}\leftarrow \mathbb{M}$
 		
 	\STATE	Make the prediction via input $\mathbb{S}$
 		
 	\STATE	$\mathcal{L}^\text{CL} \leftarrow \mathcal{L}^\text{CL}({x}^{(\mathbb{S})})$
 		
 		
 	\STATE	$\mathcal{L}^\text{CML} \leftarrow 0$
 		\FOR{$m=M-1,\ldots,1$}{
 	\STATE	Randomly erase a modality of $\mathbb{S}$ and set it as $\mathbb{T}$
 		
 	\STATE	Make the prediction via input $\mathbb{T}$
 		
 	\STATE	$\mathcal{L}^\text{CL} \leftarrow \mathcal{L}^\text{CL} + \mathcal{L}^\text{CL}({x}^{(\mathbb{T)}})$
 		
 	\STATE	\textcolor{violet}{$\mathcal{L}^\text{CML} \leftarrow \mathcal{L}^\text{CML} + \max\left(0, \text{Conf}({x}^{(\mathbb{T})})-\text{Conf}({x}^{(\mathbb{S})})  \right)$} 
}
 \ENDFOR
 	\STATE $\mathcal{L}=\frac{1}{M}\mathcal{L}^\text{CL}\textcolor{violet}{+\lambda \mathcal{L}^\text{CML}}$
 		
 	\STATE	Update the parameters of the classification model with $\mathcal{L}$}
   \ENDFOR
 	\STATE \textbf{return} the classifier $f_\text{CL}$
  \end{algorithmic}
 \end{algorithm}

\subsection{CML for Imputation-dependent Model}
For imputation-dependent method, we use MIWAE to train the reconstruction model first, then we use the reconstructed modalities to train the classifier.

\begin{algorithm}[ht]
 	\caption{CML for the imputation-dependent model}
 	 	\label{alg:recon}
    \begin{algorithmic}
 \STATE   \textbf{Given} dataset $\mathcal{D}=\left\{\{{x}_i^m\}_{m=1}^M,y_i\right\}_{i=1}^N$, reconstruction network $f_\text{re}$ and classifier $f_\text{CL}$, reconstruction loss function $\mathcal{L}^\text{re}$, Coefficient $\lambda$ of CML, epochs for training the reconstruction net ${epoch}_\text{re}$ and classifier ${epoch}_\text{CL}$
 		
 		\FOR{$e_1=1,\ldots,{epoch}_\text{re}$}{ 
 		
 	\STATE	Reconstruct the modalities via reconstruction model
 		
 	\STATE	Compute the reconstruction loss by $\mathcal{L}^\text{re}$ 
 		
 	\STATE	Update the parameters of the reconstruction model
 		}
   \ENDFOR
 		\FOR{$e_2=1,\ldots,{epoch}_\text{CL}$}{ 
 		
 	\STATE	$\mathbb{S}\leftarrow \mathbb{M}$
 		
 	\STATE	$\mathcal{L}^\text{CE} \leftarrow \mathcal{L}^\text{CE}({x}^{(\mathbb{S})})$

 \STATE	$\mathcal{L}^\text{CML} \leftarrow 0$
 		
 		\FOR{$m=M-1,\ldots,1$}{
 	\STATE	Randomly erase a modality of $\mathbb{S}$ and set it as $\mathbb{T}$
 		
 	\STATE	Reconstruct the erased modalities via reconstruction model and add them to ${x}^{(\mathbb{T})}$
 		
 	\STATE	Compute the classification loss $\mathcal{L}^\text{CE}({x}^{(\mathbb{T})})$ with Cross-Entropy loss function
 		
 	\STATE	$\mathcal{L}^\text{CE} \leftarrow \mathcal{L}^\text{CE} + \mathcal{L}^\text{CE}({x}^{(\mathbb{T})})$
 		
 	\STATE	\textcolor{violet}{$\mathcal{L}^\text{CML} \leftarrow \mathcal{L}^\text{CML} + \max\left(0, \text{Conf}({x}^{(\mathbb{T})})-\text{Conf}({x}^{(\mathbb{S})}) \right)$} 
 		}
 	\ENDFOR
 	\STATE	$\mathcal{L}=\frac{1}{M}\mathcal{L}^\text{CE}\textcolor{violet}{+\lambda \mathcal{L}^\text{CML}}$
 		
 	\STATE	Update the parameters of the classification model with $\mathcal{L}$
 		}
 	\ENDFOR
 \STATE	\textbf{return} the reconstruction model $f_\text{re}$ and classifier $f_\text{CL}$
  \end{algorithmic}
 \end{algorithm}

For reconstruction-based method, the missing modalities need to be reconstructed first, so the process can be divided into two stages.


\section{Discussion}
\subsection{Class-imbalanced} \label{sec:app-imbalance}
$\circ$ \textbf{Why the CML can still work when the training data is class-imbalanced (e.g., long-tailed)?}

\begin{figure}[ht]
    \centering
    \includegraphics[width=1.0\textwidth]{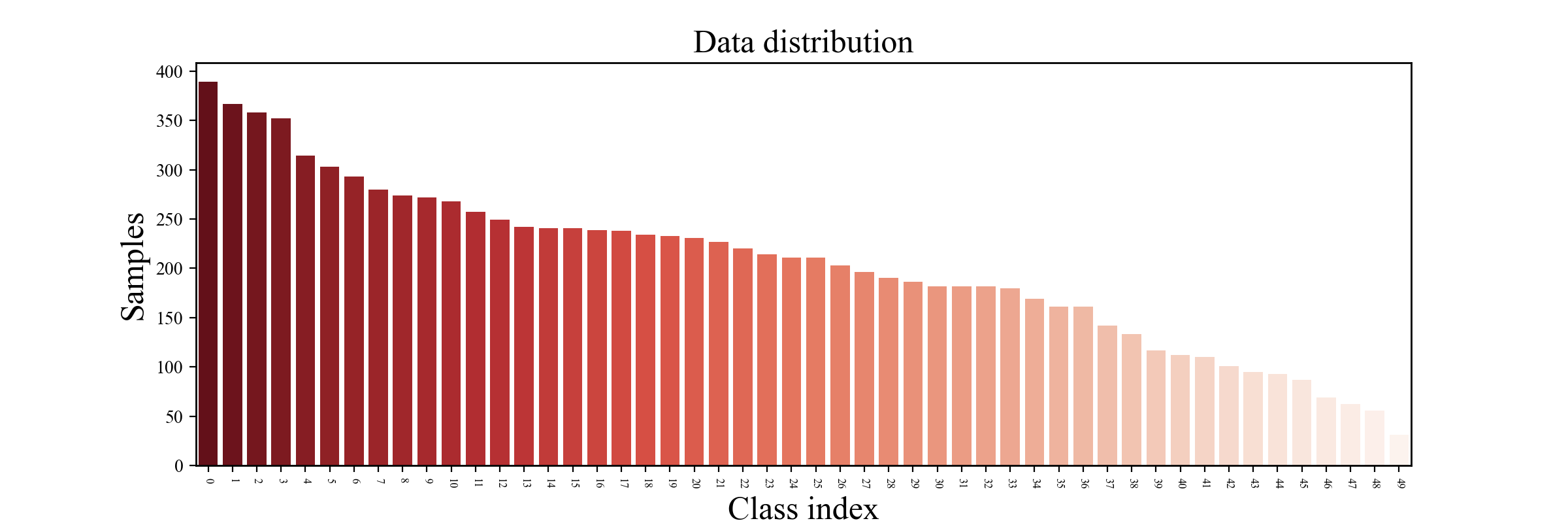}
    \caption{Illustration of data distribution of Animal dataset (the number of samples for every classes).}
    \label{fig:animal}
\end{figure}

CML can improve performance when the data for the training model is class-imbalanced since it increases the confidence of the minority classes. For a trustworthy model, the model should treat the majority and minority classes equally during the test. CML requires the model to make predictions fairly regardless of whether the majority and minority classes of the samples belong. On the contrary, the original model tends to predict lower confidence for the minority classes than the majority classes. And the improvements on the class-imbalanced dataset Animal (data distribution is shown in Fig.~\ref{fig:animal}) validate the effectiveness.

Animal is a class-imbalanced real-world dataset, the improvement shows CML can also deal with applications that suffer from class-imbalanced. The original model tends to predict lower confidence for the minority classes than the majority classes, which is unfair to minority classes. CML requires the model to make predictions fairly regardless of whether the majority and minority classes of the samples belong.



\begin{table}[th]

\begin{center}
\caption{Accuracy performance comparison of MMTM when some of the modalities is corrupted with color jitter (i.e., randomly change the brightness, contrast, saturation and hue of an image with jitter factor $\epsilon$.).}
\label{tab:mmtm-noise-jitter}
\center
\resizebox{0.94\textwidth}{!}{
\begin{tabular}{c|c|ccccc}
\toprule
\multicolumn{1}{c}{\text{Dataset}}  & \multicolumn{1}{c}{\text{Noise on}}  & \multicolumn{1}{c}{\text{CML}}  &
\text{$\mathbf{\epsilon=0.1}$}& \text{\begin{tabular}[c]{@{}c@{}}$\mathbf{\epsilon=0.2}$\\ \end{tabular}}& \text{\begin{tabular}[c]{@{}c@{}}$\mathbf{\epsilon=0.3}$\\ \end{tabular}}& \text{\begin{tabular}[c]{@{}c@{}}$\mathbf{\epsilon=0.5}$\\ \end{tabular}}
\\ \midrule
\multirow{10}{*} {\text{NYUD-2}}  
    &   \multirow{3}{*}{\text{\{1\}}}      & \xmark                                         &$65.72\pm0.70$  
       &$64.13\pm1.78$   
       &$63.79\pm1.79$
       &$60.89\pm1.21$ \\
                                                                                 &           & \cmark            
                                                         &${66.64\pm1.22}$   
                                                         &${65.41\pm0.65}$   
                                                         &${64.31\pm0.92}$ 
                                                         &${62.26\pm1.77}$               \\& & Improve & \textcolor{mycolor2}{{$~~~~~~~~\bigtriangleup 0.92 $}} & \textcolor{mycolor2}{{$~~~~~~\bigtriangleup 1.28 $}}& \textcolor{mycolor2}{{$~~~~~~\bigtriangleup 0.52 $}}  & \textcolor{mycolor2}{{$~~~~~~\bigtriangleup 1.37 $}}                     \\ \cmidrule{2-7}
      &  \multirow{3}{*}{\text{\{2\}}}       & \xmark                                                    &$61.34\pm0.98$  &$57.98\pm0.81$   &$53.98\pm2.28$   &$52.26\pm3.23$    \\
                                                                                  
                                                            &                           & \cmark                                                                                            &${62.63\pm0.60}$   &${57.89\pm1.56}$   &${54.80\pm2.90}$  &${52.57\pm3.38}$   \\& & Improve & \textcolor{mycolor2}{{$~~~~~~~~\bigtriangleup 1.29 $}} & \textcolor{mycolor1}{{$~~~~~~\bigtriangledown 0.09 $}}& \textcolor{mycolor2}{{$~~~~~~\bigtriangleup 0.82 $}}  & \textcolor{mycolor2}{{$~~~~~~\bigtriangleup 0.31 $}}                                     \\ \cmidrule{2-7}
     &  \multirow{3}{*}{\text{\{1, 2\}}}      & \xmark                                                      &$60.43\pm0.82$  &$55.17\pm0.85$  &$51.01\pm2.64$  &$41.52\pm4.01$                                         \\
                                                                                 &             & \cmark                                                       &${61.87\pm0.93}$  &${56.24\pm2.22}$   &${51.53\pm1.91}$         &${41.99\pm3.37~~}$                        \\& & Improve & \textcolor{mycolor2}{{$~~~~~~~~\bigtriangleup 1.44 $}} & \textcolor{mycolor2}{{$~~~~~~\bigtriangleup 1.07 $}}& \textcolor{mycolor2}{{$~~~~~~\bigtriangleup 0.52 $}}  &  \textcolor{mycolor2}{{$~~~~~~\bigtriangleup 0.47 $}}                                                                              \\ \midrule
\multirow{10}{*} {\text{SUN-RGBD}}  
     &  \multirow{3}{*}{\text{\{1\}}}      & \xmark                                                             &$60.72\pm0.58$  
       &$58.98\pm0.72$   
       &$57.40\pm0.75$
       &$55.68\pm0.95$ \\
                                                                                 &           & \cmark            
                                                         &${61.50\pm0.59}$   
                                                         &${59.95\pm0.17}$   
                                                         &${57.97\pm0.30}$ 
                                                         &${57.21\pm0.32}$ \\& & Improve & \textcolor{mycolor2}{{$~~~~~~~~\bigtriangleup 0.78 $}} & \textcolor{mycolor2}{{$~~~~~~\bigtriangleup 0.97 $}}& \textcolor{mycolor2}{{$~~~~~~\bigtriangleup 0.57 $}}  &  \textcolor{mycolor2}{{$~~~~~~\bigtriangleup 1.53 $}}                                      \\ \cmidrule{2-7}
      &  \multirow{3}{*}{\text{\{2\}}}       & \xmark                                                     &$60.11\pm0.24$  &$58.57\pm0.60$   &$57.46\pm0.69$  &     $55.25\pm1.05$    \\
                                                            &                       & \cmark                           &${59.90\pm0.49}$  &${58.44\pm0.75}$   &${57.25\pm0.56}$           &${55.34\pm0.87}$    \\& & Improve & \textcolor{mycolor1}{{$~~~~~~~~\bigtriangledown 0.21 $}} & \textcolor{mycolor1}{{$~~~~~~\bigtriangledown 0.13 $}}& \textcolor{mycolor1}{{$~~~~~~\bigtriangledown 0.21 $}}  &  ${-}$                                    \\ \cmidrule{2-7}
     &  \multirow{3}{*}{\text{\{1, 2\}}}      & \xmark                                                        &$58.67\pm0.42$   &$54.77\pm0.44$   &$51.66\pm0.64$        &$45.68\pm1.35$                                 \\
                                                                                 &             & \cmark                                       &${58.95\pm0.20}$
                                          &${54.73\pm0.71}$               &${51.36\pm0.66}$      &${45.99\pm1.24}$                                               \\& & Improve & \textcolor{mycolor2}{{$~~~~~~~~\bigtriangleup 0.28 $}} & ${-}$& \textcolor{mycolor1}{{$~~~~~~\bigtriangledown 0.30 $}}  &  \textcolor{mycolor2}{{$~~~~~~\bigtriangleup 0.31 $}}                                         \\   \bottomrule
\end{tabular}}
\end{center}
\end{table}

\begin{table}[th]

\begin{center}
\caption{Accuracy performance comparison of MMTM when some of the modalities is corrupted with gaussian noise (i.e., zero mean with varying variance $\epsilon$).}
\label{tab:mmtm-noise-gaussian}
\center
\resizebox{0.94\textwidth}{!}{
\begin{tabular}{c|c|ccccc}
\toprule
\multicolumn{1}{c}{\text{Dataset}}  & \multicolumn{1}{c}{\text{Noise on}}  & \multicolumn{1}{c}{\text{CML}}  &
\text{$\mathbf{\epsilon=0.1}$}& \text{\begin{tabular}[c]{@{}c@{}}$\mathbf{\epsilon=0.2}$\\ \end{tabular}}& \text{\begin{tabular}[c]{@{}c@{}}$\mathbf{\epsilon=0.3}$\\ \end{tabular}}& \text{\begin{tabular}[c]{@{}c@{}}$\mathbf{\epsilon=0.5}$\\ \end{tabular}}
\\ \midrule
\multirow{10}{*} {\text{NYUD-2}}  
    &   \multirow{3}{*}{\text{\{1\}}}      & \xmark                                         &$64.77\pm1.76$  
       &$63.03\pm1.92$   
       &$61.50\pm2.83$
       &$58.81\pm4.05$ \\
                                                                                 &           & \cmark            
                                                         &${65.26\pm1.92}$   
                                                         &${63.98\pm1.60}$   
                                                         &${62.94\pm1.97}$ 
                                                         &${59.88\pm3.03}$               \\& & Improve & \textcolor{mycolor2}{{$~~~~~~~~\bigtriangleup 1.49 $}} & \textcolor{mycolor2}{{$~~~~~~\bigtriangleup 0.95 $}}& \textcolor{mycolor2}{{$~~~~~~\bigtriangleup 1.44 $}}  & \textcolor{mycolor2}{{$~~~~~~\bigtriangleup 1.07 $}}                     \\ \cmidrule{2-7}
      &  \multirow{3}{*}{\text{\{2\}}}       & \xmark                                                    &$65.41\pm1.27$  &$62.17\pm1.76$   &$59.08\pm1.54$   &$55.75\pm2.75$    \\
                                                                                  
                                                            &                           & \cmark                                                                                            &${66.12\pm1.10}$   &${62.75\pm1.26}$   &${59.79\pm2.23}$  &${55.90\pm3.38}$   \\& & Improve & \textcolor{mycolor2}{{$~~~~~~~~\bigtriangleup 1.29 $}} & \textcolor{mycolor2}{{$~~~~~~\bigtriangleup 0.58 $}}& \textcolor{mycolor2}{{$~~~~~~\bigtriangleup 0.71 $}}  & \textcolor{mycolor2}{{$~~~~~~\bigtriangleup 0.15 $}}                                     \\ \cmidrule{2-7}
     &  \multirow{3}{*}{\text{\{1, 2\}}}      & \xmark                                                      &$61.87\pm0.82$  &$55.60\pm2.61$  &$48.62\pm4.32$  &$37.68\pm4.94$                                         \\
                                                                                 &             & \cmark                                                       &${63.12\pm1.49}$  &${57.31\pm1.58}$   &${49.51\pm2.75}$         &${37.98\pm5.21~~}$                        \\& & Improve & \textcolor{mycolor2}{{$~~~~~~~~\bigtriangleup 1.25 $}} & \textcolor{mycolor2}{{$~~~~~~\bigtriangleup 1.71 $}}& \textcolor{mycolor2}{{$~~~~~~\bigtriangleup 0.89 $}}  &  \textcolor{mycolor2}{{$~~~~~~\bigtriangleup 0.30 $}}                                                                              \\ \midrule
\multirow{10}{*} {\text{SUN-RGBD}}  
     &  \multirow{3}{*}{\text{\{1\}}}      & \xmark                                                             &$60.69\pm0.65$  
       &$58.78\pm0.95$   
       &$56.84\pm1.13$
       &$53.14\pm1.32$ \\
                                                                                 &           & \cmark            
                                                         &${61.00\pm0.32}$   
                                                         &${59.31\pm0.83}$   
                                                         &${57.47\pm0.62}$ 
                                                         &${54.77\pm1.00}$ \\& & Improve & \textcolor{mycolor2}{{$~~~~~~~~\bigtriangleup 0.31 $}} & \textcolor{mycolor2}{{$~~~~~~\bigtriangleup 0.53 $}}& \textcolor{mycolor2}{{$~~~~~~\bigtriangleup 0.63 $}}  &  \textcolor{mycolor2}{{$~~~~~~\bigtriangleup 1.63 $}}                                      \\ \cmidrule{2-7}
      &  \multirow{3}{*}{\text{\{2\}}}       & \xmark                                                     &$60.93\pm0.58$  &$59.25\pm0.71$   &$57.55\pm1.08$  &     $54.81\pm1.66$    \\
                                                            &                       & \cmark                           &${61.25\pm0.59}$  &${59.19\pm0.68}$   &${57.50\pm1.27}$           &${54.34\pm1.93}$    \\& & Improve & \textcolor{mycolor2}{{$~~~~~~~~\bigtriangleup 0.32 $}} & ${-}$ & ${-}$  &  \textcolor{mycolor1}{{$~~~~~~\bigtriangledown 0.47 $}}                                    \\ \cmidrule{2-7}
     &  \multirow{3}{*}{\text{\{1, 2\}}}      & \xmark                                                        &$59.16\pm0.88$   &$53.56\pm1.51$   &$47.22\pm2.12$        &$35.90\pm2.38$                                 \\
                                                                                 &             & \cmark                                       &${59.59\pm1.09}$
                                          &${54.14\pm0.58}$               &${47.38\pm1.47}$      &${36.30\pm2.39}$                                               \\& & Improve & \textcolor{mycolor2}{{$~~~~~~~~\bigtriangleup 0.43 $}} & \textcolor{mycolor2}{{$~~~~~~\bigtriangleup 0.58 $}}& \textcolor{mycolor2}{{$~~~~~~\bigtriangleup 0.16 $}}  &  \textcolor{mycolor2}{{$~~~~~~\bigtriangleup 0.40 $}}                                         \\   \bottomrule
\end{tabular}}
\end{center}
\end{table}

\subsection{Pair-wise Sampling}
The exact computation of the proposed loss needs to enumerate all modality set pairs (i.e., ${\mathbb{T}}$ and ${\mathbb{S}}$), which is typically computational expensive sometimes. Therefore, we introduce a strategy~\cite{moon2020Confidence,toneva2018empirical} to approximate this loss by sampling modality set pairs and find this strategy works well in practice. If the complexity of the traditional model is o(n), the complexity of our method will be o((k-1)n), where k indicates the number of modalities.

\section{CML being Deployed in Advanced Multimodal Models}\label{sec:app-advanced}

MMTM is a state-of-the-art method in multimodal classification which is selected as a representative method by \cite{wu2022Characterizing} and originally proposed by \cite{joze2020mmtm}. 
NYU Depth V2 and SUN RGB-D are two widely used multimodal datasets for RGB-D scene recognition. $\circ$~\textbf{NYUD2}: Following previous work~\cite{geo2002From}, we use a reorganized version of this dataset, which contains $1449$ samples, $10$ scene classes. $\circ$~\textbf{SUN RGB-D}~\cite{perkins2003Online}: This is a standard database of RGB-D scene recognition. Similar to previous work~\cite{geo2002From}, we also use a subset of this dataset which contains the 19 major scene categories and 9504 samples in total.
Following the author's implementation, We employ pre-trained ResNet-18 as the backbone network for MMTM. The input images are fed into depth and visual block first. Then the rgb and depth features are fused by MMTM before the final prediction. We add CML regularization to the softmax output before and after MMTM fusion process. In our experiment, the squeeze ratio of MMTM Module is set to 16. The dimensionalities of rgb and depth feature are both 512.

\begin{table}[th]

\begin{center}
\caption{VRR ($\%$) of test samples (a lower value indicates a better confidence estimation). ``\xmark'' indicates the model is not equipped with the proposed regularization ($\lambda=0$).}
\label{tab:disorder-mmtm}
\center

 \setlength{\tabcolsep}{1.4mm}
 \begin{tabular}{c|ccc}
\toprule
\multicolumn{1}{c}{\text{Method}}   & \text{CML}   & \text{NYUD-2} & \text{SUN-RGBD} \\ \midrule
\multirow{3}{*} {$\text{Type III}$}   & \xmark    &  $58.09\pm4.46$                                                    &$57.09\pm1.50$                                                       \\
                                                                                  
        & \cmark                                                                                   &$46.99\pm2.89$                                                    &$52.56\pm3.49$       \\ & Improve & \textcolor{mycolor2}{{$~~~~~~\bigtriangleup 11.10 $}} & \textcolor{mycolor2}{{$~~~~~~\bigtriangleup 4.53 $}}                                                            \\\bottomrule
\end{tabular}
\end{center}

\end{table}

\section{Related Work Details}\label{sec:app-related}
{Uncertainty estimation} provides a way for trustworthy prediction~\cite{abdar2020asurvy}. Uncertainty can be used as an indicator of whether the predictions given by models are prone to be wrong. Many uncertainty-based models have been proposed in the past decades, such as Bayesian neural networks~\cite{neal2012bayesian,mackay1992bayesian,denker1990transforming,kendall2017uncertainties}, Dropout~\cite{2017dropout}, and Deep ensembles~\cite{balaji2018ensemble,havasi2020training}. Built upon RBF networks, DUQ~\cite{van2020uncertainty} is able to identify the out-of-distribution samples, which uses distance to represent the prediction uncertainty. {Prediction confidence} is always referred to in classification models, which expects the predicted class probability to be consistent with the empirical accuracy. Models are frequently overconfident because softmax probabilities are computed with the fast-growing exponential function~\cite{hen2017Abase}, so many methods focus on smoothing the prediction probabilities distribution, such as Label smoothing~\cite{M2019when}. The recent approach employs the focal loss to calibrate the deep neural networks~\cite{Mukhoti2020Calibrating}.
A recent work~\cite{2020address} introduces True Class Probability (TCP) to ensure the low confidence for the failure predictions. Temperature scaling (TS)~\cite{guo2017On} is a well-known post-hoc confidence calibration method, which aims to re-scale the output probability by manipulating the softmax inputs, i.e., the logits.

Recently, there have been a wide range of research interests in handling missing modalities for multimodal learning, including imputation-independent methods~\cite{zhang2019cpm} and imputation-dependent methods~\cite{mattei2019miwae,wu2018multimodal}. Imputation-independent methods have no need to reconstruct the missing modalities and make classification via an uniform representation. For imputation-dependent methods (based on reconstruction), the strategy model can be split into two stages, reconstructing the missing modalities and making classification according to the reconstructed modalities. CPM-Nets~\cite{zhang2019cpm} is an advanced method which can guarantee the performance by fully exploiting all samples and all modalities to produce structured representation for interpretability, and the method has been extended and deployed into medical domain~\cite{lee2021variational}. MIWAE~\cite{mattei2019miwae} is a typical reconstruction model in multimodal classification, whose  objective is a lower bound of the likelihood of the observed data that can be tight in the limit of very large computational power.

\section{Refinement and modification following peer review}
\subsection{Underlying reason of why the confidence violates the condition}

(1) The most likely reason is the "greedy" nature of multimodal learning. Prior research~\cite{han2021greedy} has acknowledged that multimodal learning models often exhibit over-reliance on certain modalities while under-training on others, resulting in over-confidence on one input modality and an increase in confidence (statistically) when other modalities are removed.

(2) To verify this hypothesis, we assessed whether the degree of "greediness" (as defined in~\cite{han2021greedy}) and VRR are positively correlated using the Pearson correlation coefficient. We trained models with various seeds and consistently observed confidence violations in "greedy" models, as shown in the table below. Pearson correlation coefficient between VRR and Greedy~\cite{wu2022Characterizing} on SOTA method.

(3) This finding supports the notion that the proposed regularization can enhance multimodal models by mitigating their inherent greediness. Future research will explore the theoretical link between VRR and Greedy.

\subsection{Differences from traditional calibration metrics}

The proposed metric is distinct from external metrics that utilize class labels, as it is the first internal metric designed to assess calibration. The differences between external metrics and internal metrics can be analogous to clustering metrics.

(1) The proposed metric is an internal metric, while ECE and Brier score are external metrics.

(2) External metrics using class labels evaluate whether the model's confidence and accuracy are aligned from a global classification perspective. The proposed internal metric, however, is labels-free and assesses whether a model inherently meets certain criteria.

(3) We anticipate that additional internal metrics will be introduced in the future, analogous to the clustering field, and this work will benefit the community.

\subsection{Connection to unbalanced multimodal problem}

(1) The proposed method can address the problem of relying on partial modalities, as demonstrated in Table 4 and Table 5 in Appendix.

(2) The model becomes more robust when one of the modalities is corrupted, which can be considered as unbalanced multimodal problem.

(3) We evaluate the relationship between the VRR and Greedy (defined in~\cite{wu2022Characterizing} which indicates the degree of over-relying on a certain modality) by calculating the Pearson correlation coefficient according to different seeds. Pearson correlation coefficients between VRR and Greedy on SOTA method (i.e., MMTM) are 0.940 and 0.915 on NYUD2 and SUN-RGBD dataset respectively. According to empirical results, confidence violation always occurs with “greedy”.

\subsection{Analysis of loss function sampling approach}

(1) In practice, enumerating all pairs would involve permutation and combination, making it computationally expensive (detailed complexity analyses can be found in Appendix E.2).

(2) Hence, we use a sampling strategy to approximate the loss function, as demonstrated in Appendix A. The sampling approach has been widely used in various methods that encounter the same problem~\cite{toneva2018empirical,moon2020Confidence}, and has shown good approximation ability and stability.

(3) In our experiments, we introduce this sampling approach since it is widely used.

\subsection{Analysis of  hyper parameters}

(1) We choose the value of that achieves the best performance on the validation set {1, 5, 10, …, 100}.

(2) Moreover, as demonstrated in the ablation study (Fig.~4), the proposed regularization is not sensitive to the hyperparameter. Promising performance can be achieved with a mild regularization strength, indicating that the proposed regularization is not sensitive to hyperparameters and can be easily deployed in a wide range of multimodal models using CML.

\begin{table}[!t]
\begin{center}
\begin{spacing}{1.35}   
\caption{Accuracy under different $\lambda$ \label{tab:multi-lambda}}
\end{spacing}
\center
{
\setlength{\tabcolsep}{1.9mm}
\begin{tabular}{ccccccc}
\toprule
Model & Dataset &
$\mathbf{\lambda=10.0}$ & $\mathbf{\lambda=20.0}$& $\mathbf{\lambda=30.0}$& $\mathbf{\lambda=50.0}$& $\mathbf{\lambda=100.0}$
\\
\midrule
\multirow{2}{*}{\text{CPM}} &Animal &$81.83\pm2.58$ &$82.56\pm1.69$ &$82.73\pm1.64$ &$82.57\pm1.78$ &$82.30\pm2.08$
\\ &CUB &$86.67\pm4.68$ &$88.33\pm4.05$ &$86.33\pm5.49$ &$87.17\pm3.05$ &$87.17\pm3.44$\\ \midrule \multirow{2}{*}{\text{MIWAE}} &Animal &$86.91\pm0.39$ &$87.40\pm0.20$ &$87.41\pm0.38$ &$87.24\pm0.30$ &$87.32\pm0.12$\\ &CUB &$93.83\pm1.63$ &$93.50\pm1.78$ &$93.67\pm2.02$ &$97.50\pm1.33$ &$93.16\pm2.07$\\ \bottomrule
\end{tabular}}
\end{center}
\vspace{-0.1cm}
\end{table}


\end{document}